\documentclass[10pt,twocolumn,letterpaper]{article}

\usepackage[pagenumbers]{cvpr} 

\usepackage{graphicx}
\usepackage{amsmath}
\usepackage{amssymb}
\usepackage{booktabs}

\usepackage{booktabs}
\usepackage{multirow}
\usepackage{multicol}
\usepackage{arydshln}
\usepackage[misc]{ifsym}
\usepackage{appendix}

\usepackage[pagebackref,breaklinks,colorlinks]{hyperref}

\usepackage[capitalize]{cleveref}
\crefname{section}{Sec.}{Secs.}
\Crefname{section}{Section}{Sections}
\Crefname{table}{Table}{Tables}
\crefname{table}{Tab.}{Tabs.}

\def\cvprPaperID{*****} 
\def\confName{CVPR}
\def\confYear{2022}

\begin{document}

\title{
    
    Bridging CLIP and StyleGAN through Latent Alignment for Image Editing


}

\author{
Wanfeng Zheng\\
Beijing University of Posts and Telecommunications\\
{\tt\small zhengwanfeng@bupt.edu.cn}
\and
Qiang Li \thanks{Corresponding author}\\
Kuaishou Technology\\
{\tt\small liqiang03@kuaishou.com}
\and
Xiaoyan Guo \\
Kuaishou Technology\\
{\tt\small guoxiaoyan@kuaishou.com}
\and
Pengfei Wan \\
Kuaishou Technology\\
{\tt\small wanpengfei@kuaishou.com}
\and
Zhongyuan Wang \\
Kuaishou Technology\\
{\tt\small wangzhongyuan@kuaishou.com}
}





\maketitle

\begin{abstract}
Text-driven image manipulation is developed since the vision-language model (CLIP) has been proposed. 
Previous work has adopted CLIP to design a text-image consistency-based objective to address this issue.
However, these methods require either test-time optimization or image feature cluster analysis for single-mode manipulation direction.
In this paper, we manage to achieve inference-time optimization-free diverse manipulation direction mining by bridging CLIP and StyleGAN through Latent Alignment (CSLA).
More specifically, our efforts consist of three parts: 1) a data-free training strategy to train latent mappers to bridge the latent space of CLIP and StyleGAN; 2) for more precise mapping, temporal relative consistency is proposed to address the knowledge distribution bias problem among different latent spaces; 3) to refine the mapped latent in $s$ space, adaptive style mixing is also proposed. 
With this mapping scheme, we can achieve GAN inversion, text-to-image generation and text-driven image manipulation.
Qualitative and quantitative comparisons are made to demonstrate the effectiveness of our method.

\end{abstract}

\begin{figure}[t]
  \centering

  \includegraphics[width=0.9\linewidth]{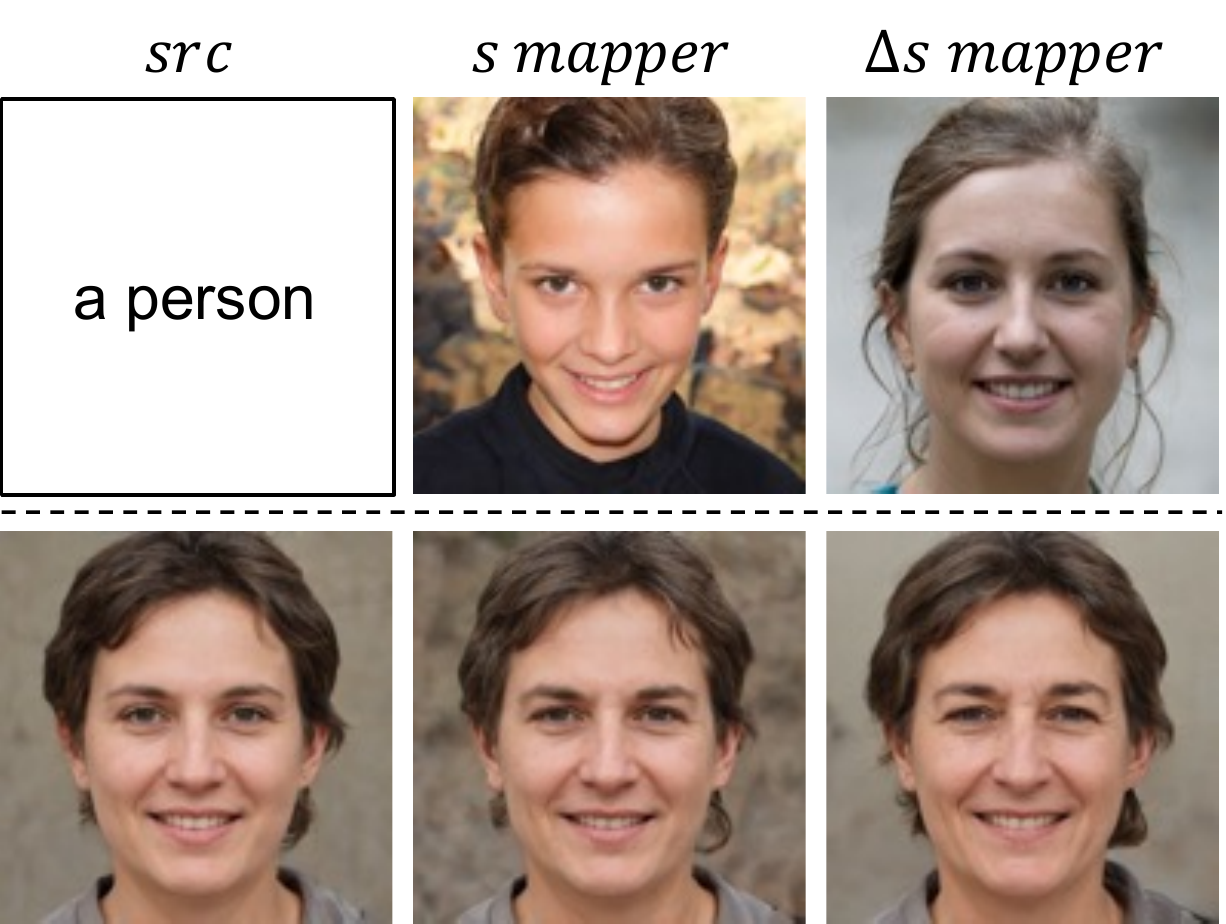}

  \caption{
    Demonstration of knowledge bias problem. 
    Images in the first row are generated from ``a person", which represents the face knowledge distribution center in CLIP-space logically. 
    It is obvious that the generated image of $\Delta s$ mapper is closer to the knowledge distribution center corresponding face of StyleGAN than $s$ mapper, which indicates the former is better aligned. 
    Images in the second row are text-driven manipulated on age. 
    And the manipulation result of $s$ mapper looks like a man rather than a woman in the source image, which means the manipulation direction is perturbed by the distribution bias.
  }
  \label{fig:bias}
\end{figure}

\section{Introduction}
\label{sec:intro}

Generative adversarial networks \cite{gan} are designed for image synthesize. 
Among several GAN variants, StyleGAN \cite{stylegan,stylegan2} has shown its superior properties on several generative tasks such as human face generation or animal face generation. 
And the generative prior of StyleGAN has been utilized in many downstream tasks \cite{jojogan,label4free,styleclip,psp,image2stylegan}. 
StyleGAN-based image manipulation is one of the most important subtopics, which indicates editing a realistic or synthesized image on its $w$, $w+$ or $s$ latent.

Due to the progress of vision-language model (CLIP) \cite{clip}, image and text data can be embedded into one shared latent space, which has enabled many text-image cross-modal approaches.
StyleCLIP \cite{styleclip} has proposed three different methods based on the text-image consistency for text-driven image manipulation: 1) latent optimization edits one image by optimizing its latent feature in $w+$ space; 2) latent mapper is trained for a certain manipulation direction and can generate residual for $w+$ latent; 3) global directions mining can explore manipulation direction in $s$ space via a statistical strategy. 
These methods suffer from lack of inference-time efficiency, because extra costs are inevitable for editing on a certain image or mining a novel manipulation direction.

\begin{figure*}[t]
  \centering
  \includegraphics[width=\linewidth]{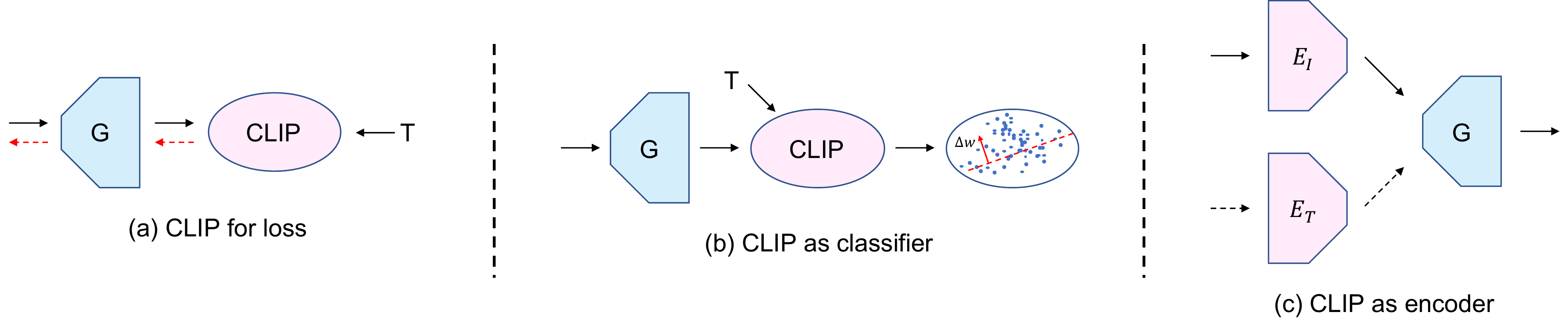}

  \caption{
    Demonstration of three major applications of CLIP.
    Red dotted line represents the data flow of backward propagation.
    Black dotted line means ``inference-only''.
    As shown in figure, the first group of works adopts CLIP to consist of their loss function.
    Another group of works analysis $\Delta w$ by zero-shot classification with CLIP.
    Besides, some works use CLIP as the image encoder and replace the image embedding with text embedding as inference time to achieve language-free training and text-driven manipulation.
  }
  \label{fig:difference}
\end{figure*}



To address this issue, we have proposed our method CSLA by bridging CLIP and StyleGAN through latent alignment. 
Specifically, we have adopted mappers to map latents from CLIP-space to $w$-space and $s$-space. 
However, there is a knowledge distribution bias problem between CLIP and StyleGAN.
These two models have knowledge divergence about the average object (e.g. the average human face), which can result in bad cases of manipulation as shown in \cref{fig:bias}.

To solve this problem and achieve more precise mapping, instead of directly mapping CLIP-latent $f_{I}$ to $w$ latent $w$ and $s$ latent $s$, we have mapped the manipulation residuals $\Delta f_{I} $ to $\Delta w$ and $\Delta s$.
Nevertheless, it is obvious to select the average latent as the manipulation center in $w$ and $s$ space, but the selection of the manipulation center in CLIP-space can greatly influence the manipulation result. 
Therefore we have discussed the effect of different manipulation center selections and proposed a temporal relative manipulation center selection strategy for training.

Despite the well-alignment of the manipulation center and manipulation direction, mapping from CLIP-space to $s$ space still has biases. 
Inspired by style mixing, we manage to design a learnable adaptive style mixing module to achieve self-modulation in $s$ space, which can generate modulating signals to refine $\Delta s$.

After training, the image encoder $E_{I}$ of CLIP can be used for GAN inversion. 
The text encoder $E_{T}$ of CLIP can be embed texts into CLIP-space as $f_{T}$. 
For the CLIP-space is shared with $f_{I}$ and $f_{T}$, $f_{T}$ can also be mapped into $w$ space or $s$ space for text-conditioned image generation. 
Besides, by given of two sentences, we can compute the manipulation direction with $f_{T_{\text{src}}}$ and $f_{T_{\text{trg}}}$.

Experiments have produced qualitative and quantitative results that can verify our method. 
Our contributions are summarized as follows:
\begin{itemize}
    \item A data-free training strategy named CSLA is proposed for arbitrary text-driven manipulation without extra time consumption at inference-time.
    
    \item We have noticed the knowledge distribution bias problem between CLIP and StyleGAN and solved this problem by mapping the latent residual.
    
    \item We have proposed temporal relative consistency and adaptive style mixing to refine mapped $\Delta s$.
    
    

\end{itemize}

\section{Related work}
\subsection{Vision-Language Model}
In recent years, vision-language pre-training \cite{clip, coop, cocop, flava, denseclip, clip-adapter} has become an attractive novel research topic. 
As the most significant study, CLIP \cite{clip} has achieved cross-modal joint representation between text and image data by training among 400 million text-image pairs from the Internet. 
With its prior text-image consistency knowledge, it can achieve SOTA on zero-shot classification. 
And its prior knowledge also promotes other text-image cross-modal tasks. 
In this paper, we employ CLIP as an image encoder for latent alignment training.

\begin{figure*}[t]
  \centering
  \includegraphics[width=\linewidth]{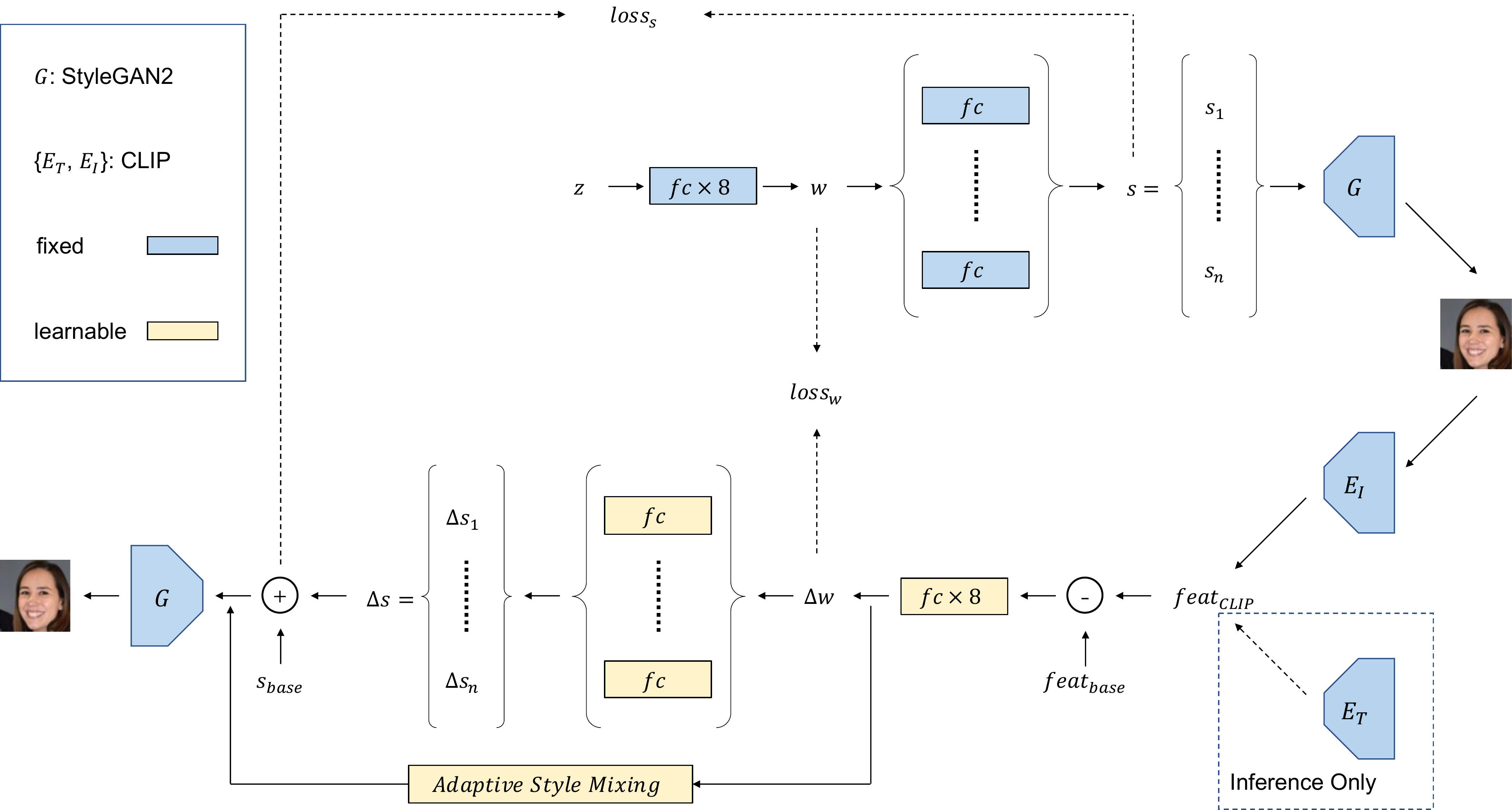}

  \caption{
    Demonstration of the training process of our method. 
    Synthesized images with their corresponding latents in $w$ and $s$ space are sampled by a pre-trained StyleGAN.
    Then the image is passed through the image encoder of CLIP and subtracted by the base feature in CLIP-space.
    Later, the residual CLIP feature is mapped into $\Delta w$ and $\Delta s=\{s_{1}, s_{2}...s_{n}\}$ which has enabled the consistency objectives to train these align mappers.
    During inference time, the image encoder of CLIP is replaced by the text encoder. 
    Therefore, we can achieve text-driven manipulation direction mining.
  }
  \label{fig:method}
\end{figure*}

\subsection{Zero-Shot Text-to-Image Generation}
Zero-shot text-to-image generation is a conditional generation task with reference text.
To achieve zero-shot text-to-image generation, DALL-E \cite{dalle} aims to train a transformer \cite{transformer} to autoregressively model the text and image tokens as a single stream of data. 
Beyond zero-shot generation, CogView \cite{cogview} has further investigated the potential of itself on diverse downstream tasks, such as style learning, super-resolution, etc.
CogView2 has made attempts for faster generation via hierarchical transformer architecture.
Different form those transformer-based emthods, DiffusionCLIP \cite{diffusionclip}, GLIDE \cite{glide} and DALL-E2 \cite{dalle2} has introduced diffusion model \cite{diffusion, diffusion2, ddpm} with CLIP \cite{clip} guidance and achieved better performance. 
From then on, many recent works \cite{imagen, vqgan-clip, clip-gen, dreambooth} have greatly developed this research direction.
Most of these works suffer from expensive training consumption due to their billions of parameters. 
Meanwhile, our method with about only 8M learnable parameters can achieve text-to-image generation via a much faster training process rather than these methods on a single 2080Ti.

\subsection{StyleGAN-based Image Manipulation}
StyleGAN has three editable latent spaces ($w$, $w+$, $s$) and high-fidelity generative ability.
Therefore, some approaches manage to achieve image manipulation through an inversion-editing manner.
pSp \cite{psp} has proposed a pyramid encoder to inverse an image for RGB space to $w+$ space.
Instead of learning inversion from image to $w+$ latent, e4e \cite{e4e} encodes an images into $w$ and a group of residuals to produce $w+$ latent.
Besides, ReStyle \cite{restyle} iteratively inverse an image into $w+$ space by addition on the predicted $w+$ residual and achieved SOTA performance.
To manipulate the inversed image latent, most of the existing non-CLIP manipulation direction mining methods are working through a clustering way, e.g. InterFaceGAN \cite{interfacegan}. 
Our method manipulates images with two sentences instead of two sets of images.

\subsection{Combination of CLIP and StyleGAN}
StyleCLIP \cite{styleclip} is the most influential CLIP-driven StyleGAN manipulation approach by using CLIP to consist its objective.
Following StyleCLIP, StyleMC \cite{stylemc} has trained mappers to produce $\Delta s$ for single-mode StyleGAN manipulation.
And StyleGAN-NADA \cite{stylegan-nada} finetunes an generator with text guidance.
Essence Transfer \cite{essence} has also adopted CLIP for its objective to achieve image-based essence transfer by optimization on latents of StyleGAN.
Another work Counterfactual \cite{counterfactual} has focused on counterfactual manipulation by learning to produce manipulation residual on $w$ latents.
Meanwhile, ContraCLIP \cite{contraclip} has made attempts to address the problem of discovering non-linear interpretable paths in the latent space of pre-trained GANs in a model-agnostic manner.
Different from these works, CLIP2StyleGAN \cite{clip2stylegan} has adopted CLIP as a classifier to tag a large set of real or synthesized images. Thus they can better disentangle the $w$ space to achieve exact manipulation.
Besides, some other works like HairCLIP \cite{hairclip} have adopted CLIP as their image encoder while our method has also utilized CLIP in this way as shown in \cref{fig:difference}.
Among this kind of methods, AnyFace \cite{anyface} is the most similar to our method.
But it requires paired language-image training data, thus can not support manipulation in arbitrary data domains.
Meanwhile, our method combines CLIP and StyleGAN in an unsupervised and training data-free manner so that we can achieve diverse text-driven image manipulation more efficiently. 
A concurrent work CLIP2Latent \cite{clip2latent} also developed latent mapping for text-to-image generation.
There are several distinctions: 
1) they adopt diffusion model for latent mapping while ours use MLP which is faster;
2) their method maps to $w$ latent while ours maps to $\Delta s$ latent for better disentangled manipulation;
3) we propose temporal relative consistency and adaptive style mixing to refine the mapping.



\begin{figure}[t]
  \centering
  \includegraphics[width=\linewidth]{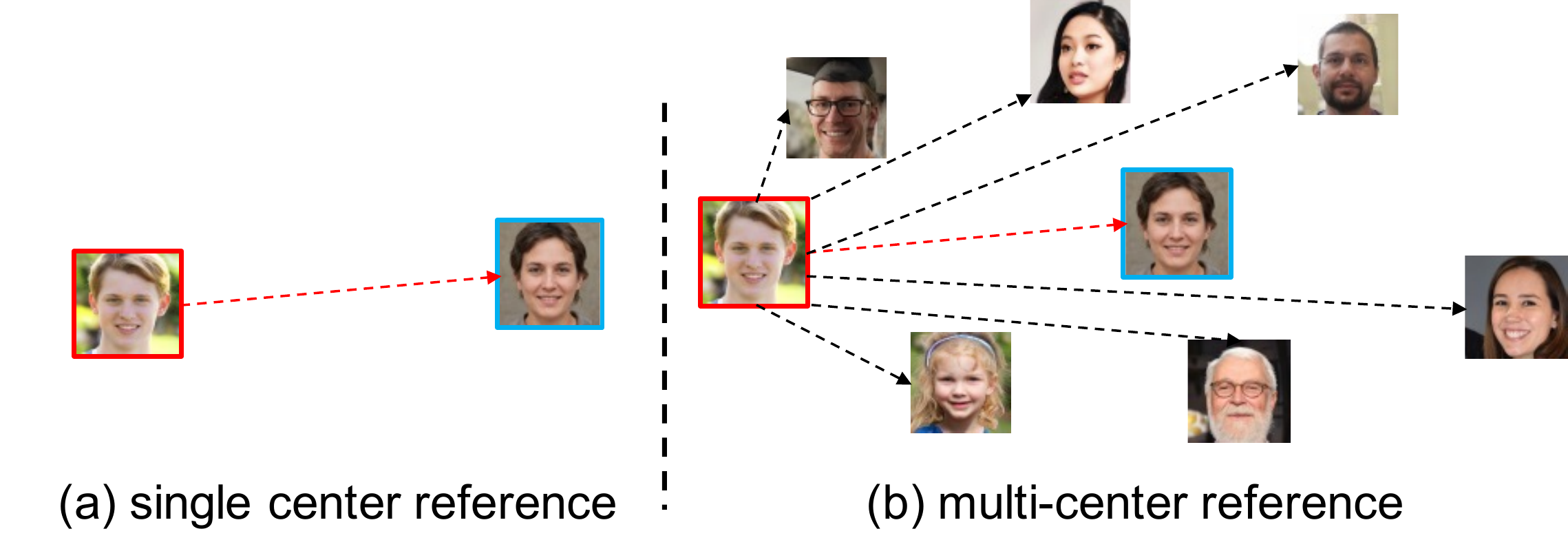}

  \caption{
    Comparison of single center reference (left) and multi-center reference (right).
    Rather than single center, multi-center has more references, which leads to more precise latent orientation in $w$ space.
    Therefore, multi-center training strategy can result in preferable latent alignment.
  }
  \label{fig:trc}
\end{figure}

\section{Method}
In this section, the description of our method has been divided into five parts: 1) the overall architecture of our method; 2) the selection of manipulation center in CLIP-space and our proposed temporal relative consistency; 3) the adaptive style mixing strategy; 4) the objective of our method; 5) how our method works at inference time.

\subsection{Overall Architecture of Latent Alignment}
The overall architecture of our method has been shown in \cref{fig:method}.
Our major purpose is to train mappers to align latent spaces. 
To efficiently achieve that, we manage this training process through a latent distillation strategy. 
During training, we sample $w$ and $s$ latents with their corresponding RGB image $I$ by a pre-trained StyleGAN.
\begin{equation}
    w, \, s, \, I = G(z), \, z \sim N(0,1).
  \label{eq:stylegan2latent}
\end{equation}

Then the image is encoded to a CLIP latent to compute residual latent $\Delta f_\text{CLIP}$ with preset CLIP-space manipulation center ${f_\text{base}}$, which is fed into full connected (\text{FC}) mappers to compute $\Delta w$ and $\Delta s$.
\begin{equation}
        \begin{aligned}
            &\Delta{f_\text{CLIP}} = E_{I}(I) - {f_\text{base}}, \\
            &\Delta{w} = \text{FC}_{w}(\Delta{f_\text{CLIP}}), \\
            &\Delta{s} = \{{\text{FC}_{s_{i}}(\Delta{w})}\} , \, i \in \{1,2...n\}.
        \end{aligned}
  \label{eq:dw_and_ds}
\end{equation}

While $\Delta w_\text{trg}$ and $\Delta s_\text{trg}$ can be computed by subtraction between original $w, s$ and the preset manipulation center latent $w_\text{base}, s_\text{base}$, our mappers can be trained by forcing $\Delta w$ and $\Delta s$ to fit $\Delta w_\text{trg}$ and $s_\text{trg}$. 
\begin{equation}
        \begin{aligned}
            \Delta{w_\text{trg}} &= w - w_\text{base}, \\
            \Delta{s_\text{trg}} &= \{s_{i} - s_{base_{i}} \}, \, i \in \{1,2...n\} . \\
        \end{aligned}
  \label{eq:trg}
\end{equation}

After all, the StyleGAN is forward propagated once for each iteration, which can save about $50\%$ cost than the distillation in RGB space. 

The network architecture of our mappers is the same as StyleGAN because its effectiveness is proven and we consider it is not necessary for extra incremental designs for the mapper.
And we have removed the pixel normalization to avoid information loss of latent norm.

\begin{figure}[t]
  \centering
  \includegraphics[width=\linewidth]{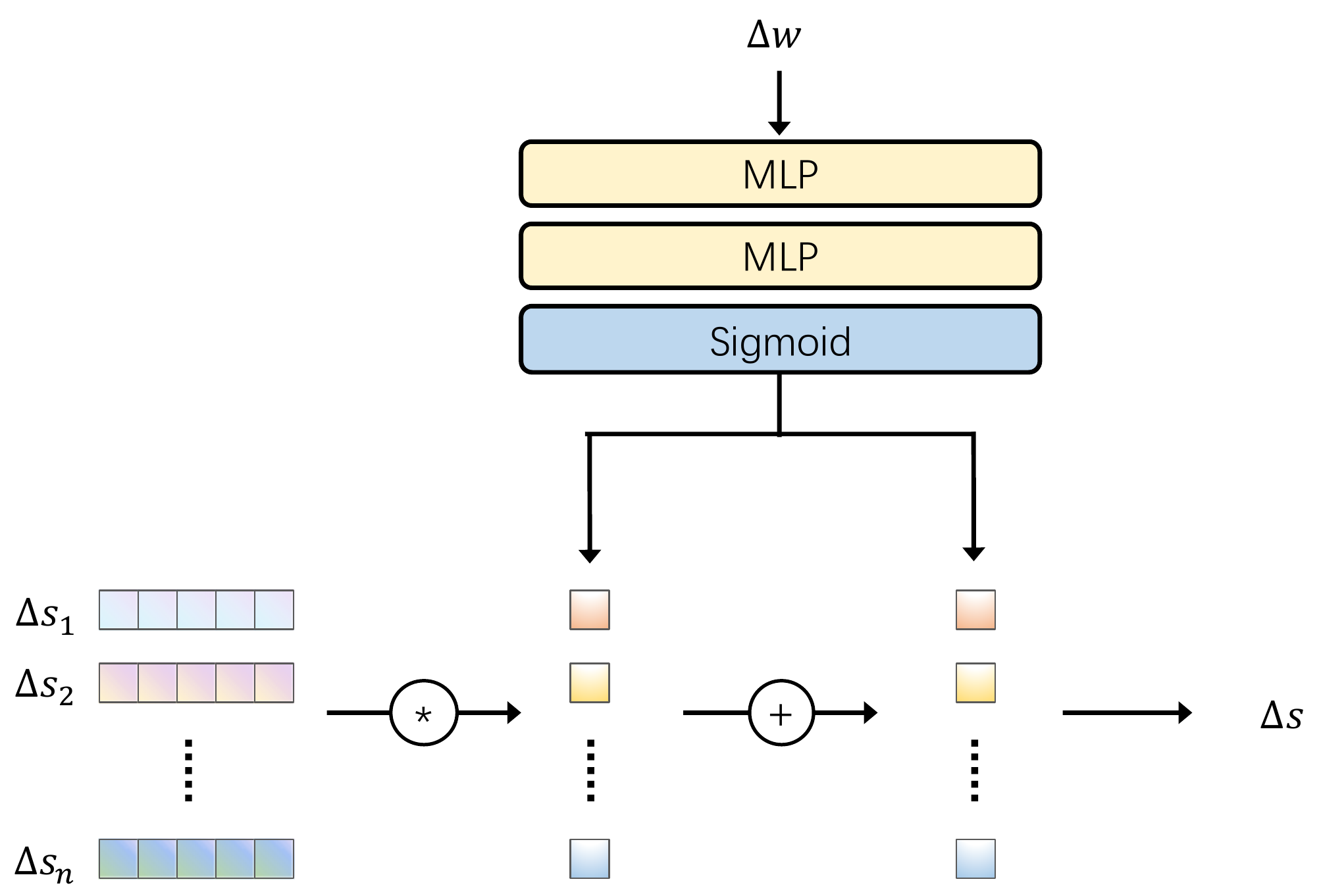}
  \caption{
    The architecture of proposed adaptive style mixing. 
    The tiny network produces two modification elements to modify each $\Delta s$ latent.
  }
  \label{fig:asm}
\end{figure}

\begin{figure*}[t]
  \centering
  \includegraphics[width=\linewidth]{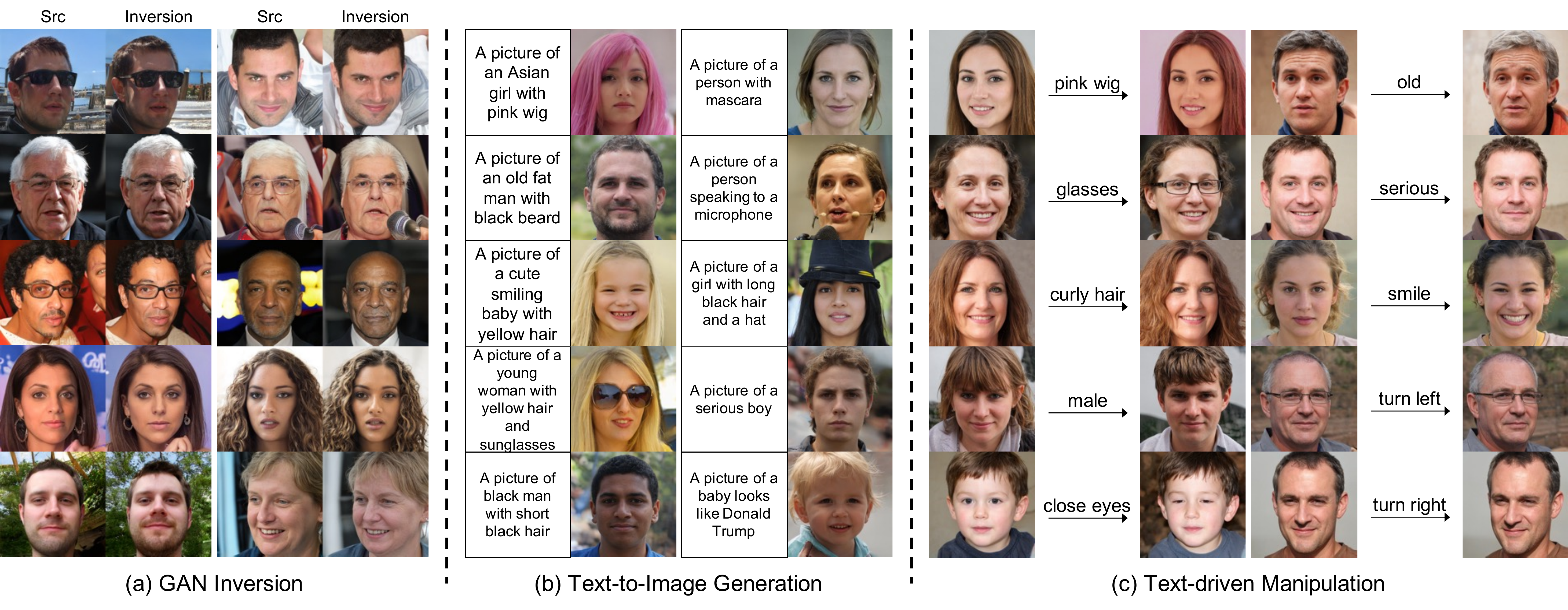}

  \caption{
    Results produced by our methods. 
    From left to right: a) GAN inversion; b) text-to-image generation; c) text-driven image manipulation.
    Our method has shown its competitiveness in text-driven image manipulation for it can not only edit the appearance or expression of the human face but also change the pose of the head.
  }
  \label{fig:demo}
\end{figure*}

\subsection{Temporal Relative Consistency}
To address the knowledge distribution bias problem discussed in \cref{sec:intro}, we manage to map latent residuals. 
Therefore, the selection of manipulation centers $f_\text{base}, w_\text{base}$ and $s_\text{base}$ are crucial for latent alignment. 
It is reasonable to select the $w_{avg}$ and its corresponding $s_{avg}$ for $w$ space and $s$ space. 
The only thing questionable is the choice of CLIP-space.
We have made attempts for at least four selections: 1) average center $\hat{f_\text{CLIP}}$ encoded from $\hat{I}$ corresponding to the average latent $\hat{w}$ of StyleGAN; 2) text center embedded from ``a picture of [class]", where [class] is the StyleGAN's specialty; 3) exponential moving average (EMA) center computed during training; 4) learnable manipulation center.

Having made a series of attempts, we have found the average center outperforms the other three configurations.
The text center lacks stability because it can be influenced by the original text description.
It is mathematical explicable that the EMA center has the same performance as the average center. 
And the learnable center can be considered as the element-wise bias on latents, which has performed worse than the average center.

To further refine alignment, we have proposed temporal relative consistency as an extension of the average center.
More specifically, in addition to the major manipulation center, we have employed two dynamic queues to keep several previous sampled $f_\text{CLIP}$ and $w$ latents.
Therefore, we can compute $\Delta f_\text{CLIP}$ and $\Delta w$ with randomly sampled multi-center instead of the single average center as shown in \cref{fig:trc}.

The effectiveness of the temporal relative consistency can be explained by implicitly expanded batch size and data augmentation.
Although we still sample one latent per iteration, with temporal relative consistency, the input batch size of $\text{FC}_{w}$ is equal to the size of queues, which has been fixed to 4096 in practice.
For the batch size expansion is made on latent-level, the extra computation cost is limited to only about 8G MACs, which has hardly affected training time consumption.
Besides, the multi-center queue is dynamic and temporal updated so that the $\Delta f_\text{CLIP}$ is strongly augmented for better convergence.

\subsection{Adaptive Style Mixing }
According to the architecture of StyleGAN \cite{stylegan}, $s$ latents in different resolution levels can be mixed and result in mixed RGB images.
This operation is named style mixing.
By using style mixing, the manipulation latent in $s$ space can be more precisely disentangled \cite{clip2stylegan}.
Inspired by this, we manage to involve style mixing in a learnable manner.
Therefore, we have proposed adaptive style mixing (ASM).

With an exemplar of StyleGAN trained on $1024\times 1024$ resolution, there are 26 latent vectors in $s$ space in total.
We design a tiny network with two MLP layers followed by a Sigmoid function with $\Delta w$ as its input to produce modulating signals as shown in \cref{fig:asm}.
The tiny network produces affine signals to control the magnitude and adjust the bias of each $s$ latent.

\begin{equation}
    \begin{aligned}
        \alpha, \, \beta = \text{ASM}(\Delta w), \, 
                    \alpha, \, \beta \in \mathcal{R}^{ n}, \\
        \Delta s_{i} = \alpha_{i} * \Delta s_{i} + \beta_{i} , 
                                    \, i \in \{1,2..n\}.
    \end{aligned} 
  \label{eq:adamix}
\end{equation}

\subsection{Objective}
The overall objective of our method consists of two parts. 
They supervise the alignment training in $w$ and $s$ space respectively.
\begin{equation}
        \begin{aligned}
            \mathcal{L}_{\text{total}} = \mathcal{L}_{w} + \lambda_{s} * \mathcal{L}_{s}.
        \end{aligned} 
  \label{eq:loss_total}
\end{equation}

We manage to supervise consistency on both the absolute value of the latent residual and the direction of the latent residual.
The direction consistency is to force maximize the cosine similarity between vectors,
\begin{equation}
        \begin{aligned}
            \mathcal{L}_{\text{dir}}(x, y) = 1 - \frac{x*y}{|x|*|y|}.
        \end{aligned} 
  \label{eq:loss_diretion}
\end{equation}

Then, we can represent objective on $\Delta w$ and $\Delta s$ as:
\begin{equation}
    \begin{aligned}
       \mathcal{L}_{w} = & |\Delta w - \Delta w_{\text{trg}}| + 
                \lambda_{w_{\text{dir}}} \mathcal{L}_{\text{dir}}(\Delta w, \Delta w_{\text{trg}}), \\
      \mathcal{L}_{s} = & \frac{1}{n} \sum_{i=1}^{n} \left[
      |\Delta s_{i} - \Delta s_{\text{trg}_{i}}| +
      \lambda_{s_{\text{dir}}} \mathcal{L}_{\text{dir}}(\Delta s_{i}, \Delta s_{\text{trg}_{i}}) \right].
    \end{aligned} 
  \label{eq:loss_w_s}
\end{equation}

Here, $n$ represents the number of latents in $s$ space. In practice, we have fixed $\lambda_{w_{\text{dir}}}$, $\lambda_{s}$ and $\lambda_{s_{\text{dir}}}$ to 10, 10 and 1.

\begin{table*}[t]
\scriptsize
  \centering

  \begin{tabular}{c|c|c:c:c|c:c:c}
    \toprule
    
    \textbf{Role of CLIP} &\textbf{Method} & \textbf{pre-proc.} 
    &\textbf{training time} &\textbf{infer. time}
    &\textbf{extra paired data} &\textbf{latent space}  
    &\textbf{manipulation mode}   \\

    \midrule
    
    \multirow{4}{*}[-0.5ex]{\textbf{Loss}} 
            & StyleCLIP-optim. \cite{styleclip} & $-$ & $-$ & 98 sec & no & $w+$ & arbitrary \\
    & StyleCLIP-mapper & $-$ & $10-12$h & 75ms & no & $w+$ & single \\
    & StyleCLIP-global dir. & 4h & $-$ & 72ms & no & $s$ & single \\
    & StyleMC \cite{stylemc} & $-$ & 5s &65ms & no & $s$ & single \\

    \hdashline 
    {\textbf{Classifier}} &  CLIP2StyleGAN \cite{clip2stylegan}
        & unknown & $-$ &$\text{30ms}^{*}$ & no & $w$ & multiple \\
    
    \hdashline
    \multirow{2}{*}[-0.5ex]{\textbf{Encoder}}
        & AnyFace \cite{anyface} & $-$ & unknown & unknown & yes & $w+$ & arbitrary \\  
        
    & Ours & $-$ & 145h & 59ms & no & $s$ & arbitrary \\  

    \bottomrule
    
    \end{tabular}
        \caption{ 
    Comparison of settings for different methods combined of CLIP and StyleGAN.
    The inference time of CLIP2StyleGAN marked by $^{*}$ is our reproduction with their open-source code.
    ``$-$'' means that the item is not necessary and should be remained blank.
    ``unknown'' means that the item is unable to reproduce due to a lack of source code or time consumption.
    The configuration ``manipulation mode'' has three types:
    ``single'' means the method only supports one manipulation mode after training or pre-processing;
    ``multiple'' means that CLIP2StyleGAN can support different but only a few manipulation modes;
    ``arbitrary'' means the method can support different manipulation modes for each inference without limitation.
  }
  \label{tab:settings}
\end{table*}

\subsection{Manipulation}
Before manipulation on latents, GAN inversion is required. 
Previous works \cite{anyface} make inversion by employing state-of-the-art GAN inversion methods.
Our latent alignment method can not only use other GAN inversion methods, but also support inversion for general image cases as a by-product via the image encoder ($E_{I}$) of CLIP.
Though the inversion performance has a deviation from the source image especially on corner cases due to most of the parameters are fixed to keep prior knowledge, it is still a convenient and efficient option.
After inversion, $w$ and $s$ latent of the source image are gained.

By giving of two sentence $T_{\text{src}}$ and $T_{\text{trg}}$, $\Delta f_{\text{src}}$ and $\Delta f_{\text{trg}}$  is computed with the major manipulation center:
\begin{equation}
    \begin{aligned}
    \Delta f_{\text{src}} &= E_{T}(T_{\text{src}}) - f_{\text{base}} ,\\
    \Delta f_{\text{trg}} &= E_{T}(T_{\text{trg}}) - f_{\text{base}}.
    \end{aligned}
  \label{eq:dt}
\end{equation}

Then we can map $\Delta T_{\text{src}}$ and  $\Delta T_{\text{trg}}$ into $w$ and $s$ space.
The manipulation direction $\Delta w$ and $\Delta s$ are computed by:
\begin{gather}
            \Delta w_{\text{src}}, \, \Delta w_{\text{trg}} = 
                    \text{FC}_{w}(\Delta f_{\text{src}}), 
                    \, \text{FC}_{w}(\Delta f_{\text{trg}}) , \\
            \Delta w = \Delta w_{\text{trg}} - \Delta w_{\text{src}}, \\
            \Delta s = \{{\text{FC}_{s_{i}}(\Delta{w_{\text{trg}_{i}}})} -
                        {\text{FC}_{s_{i}}(\Delta{w_{\text{src}_{i}}})}\} ,
                        \, i \in \{1,2...n\}.
      \label{eq:dir}
\end{gather}

With $\Delta w$ and $\Delta s$, we can achieve both $w$ and $s$ space manipulation.
But we have found that manipulation in $s$ space is more flexible and accurate because the selection of layers to modify has enabled more precise manipulation direction disentanglement.
Besides the general manipulations on appearance, we can also achieve manipulation on pose or expression as shown in \cref{fig:demo}.

\section{Experiment}
\subsection{Implementation Details} 
Our work is developed based on the open-source code of CLIP and StyleGAN \cite{stylegan-rosi}.
Experiments are made by using their pre-trained models.
While the CLIP model is unique, StyleGAN models are trained on different datasets including the human face dataset FFHQ \cite{stylegan} and cat, car, church, horse datasets from LSUN \cite{lsun}.
Image resolution is fixed to $1024 \times 1024 $ for FFHQ, $ 512 \times 512 $ for animal face and LSUN car, $ 256 \times 256 $ for the others.
During training, the batch size is set to 1.
The learning rate is initialized to 1e-2 in the beginning, then reduced by the Poly scheme.
Adam \cite{adam} optimizer is employed to update the weights of the network.
The whole training process lasts for 5e6 iterations and spends about 6 days on a single NVIDIA GeForce RTX 2080Ti.
The whole framework is implemented on Pytorch. 

Besides, we have also integrated our method on StyleGAN3 \cite{stylegan3} and StyleGAN-XL \cite{styleganxl} for experiments on animal face dataset AFHQ \cite{stargan,stargan2} and ImageNet \cite{imagenet}.
Most of the demonstration and comparison figures in this paper consist of materials from FFHQ.
Demonstrations for other datasets are in our supplementary materials for page limitation.
These results have shown our method can also manipulate motion, background and grayscale.


\begin{figure}[t]
  \centering
  \includegraphics[width=\linewidth]{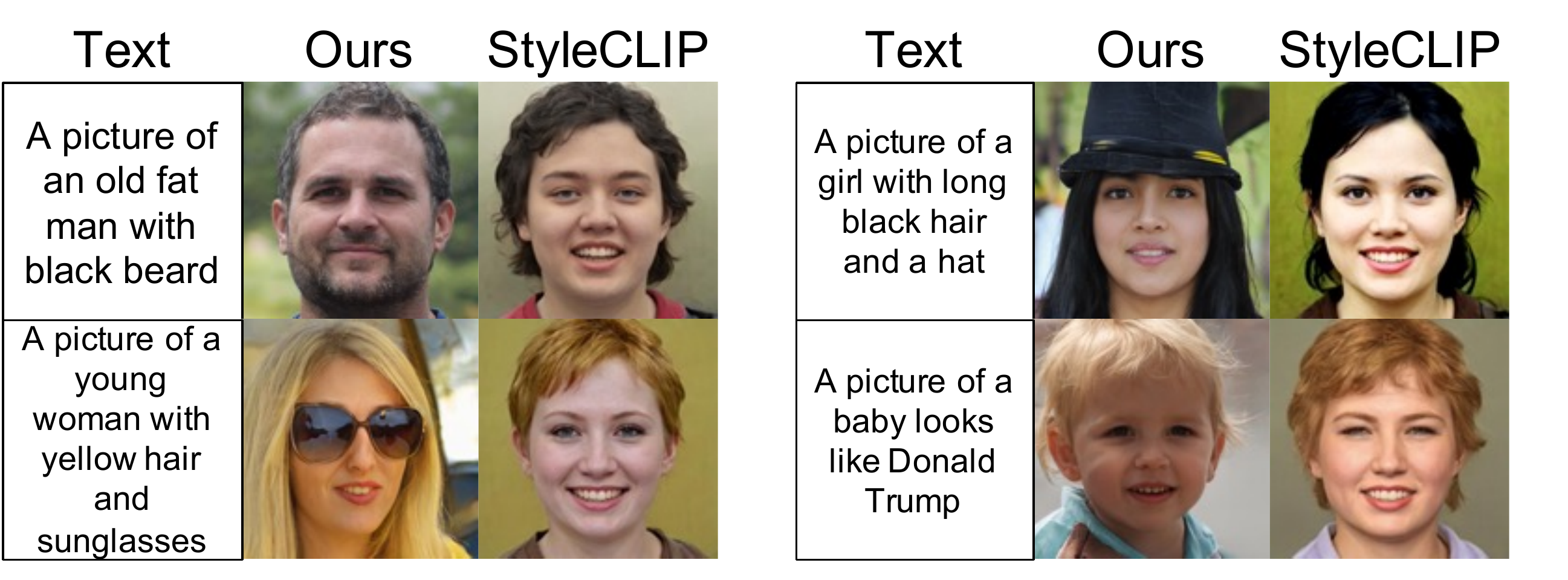}

  \caption{
    Comparison of text-to-image generation between StyleCLIP and our method.
  }
  \label{fig:styleclip_generation}
\end{figure}

\begin{figure*}[t]
  \centering
  \includegraphics[width=\linewidth]{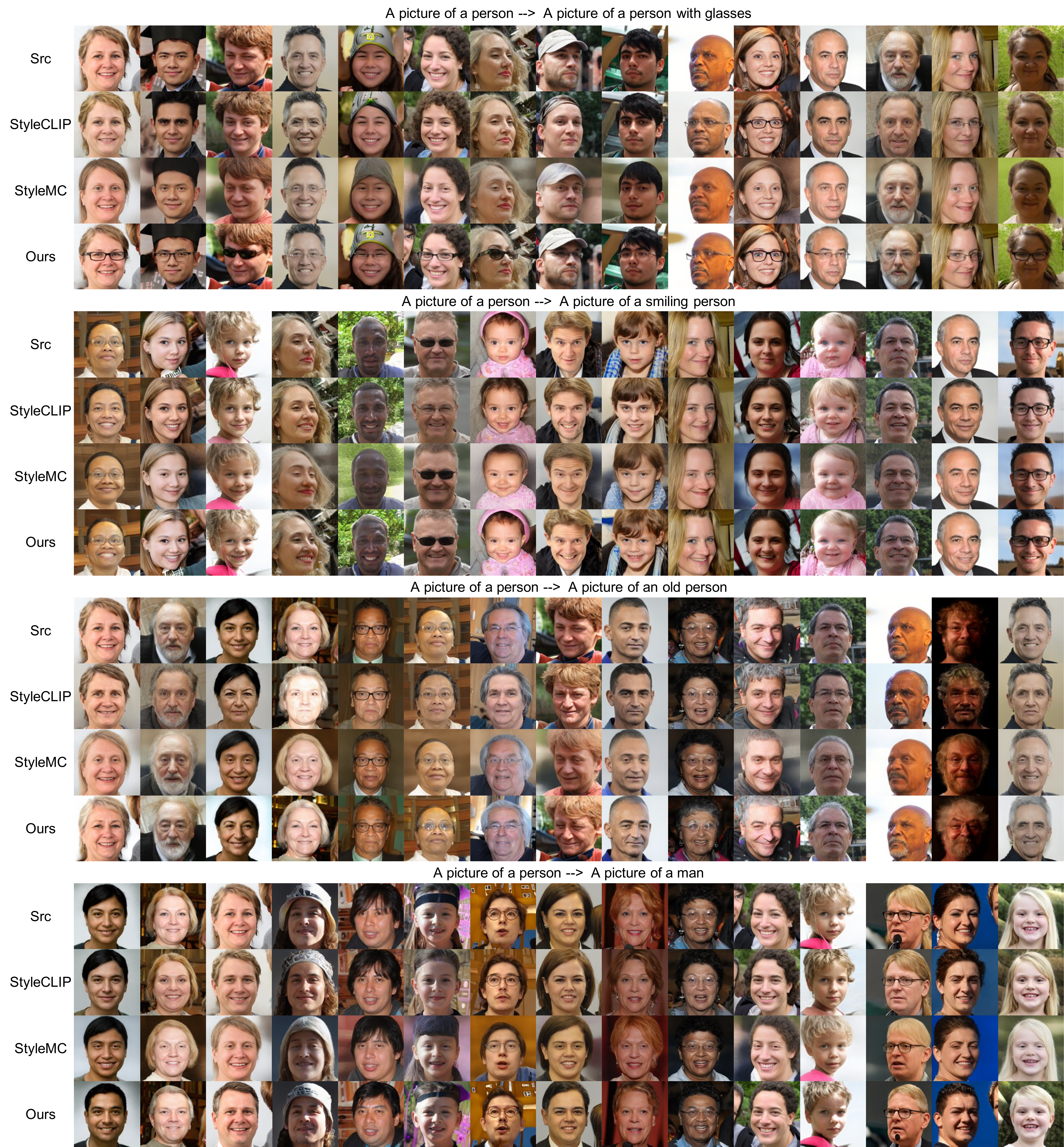}

  \caption{
    Qualitative comparison with other state-of-the-art methods.
    From top to bottom, there are four groups of manipulation results representing ``decoration'', ``expression'', ``age'' and ``gender'' respectively.
    Rather than other methods, our method can produce manipulation results better aligned with the target text prompt.
  }
  \label{fig:compare}
\end{figure*}

\begin{figure*}[t]
  \centering
  \includegraphics[width=\linewidth]{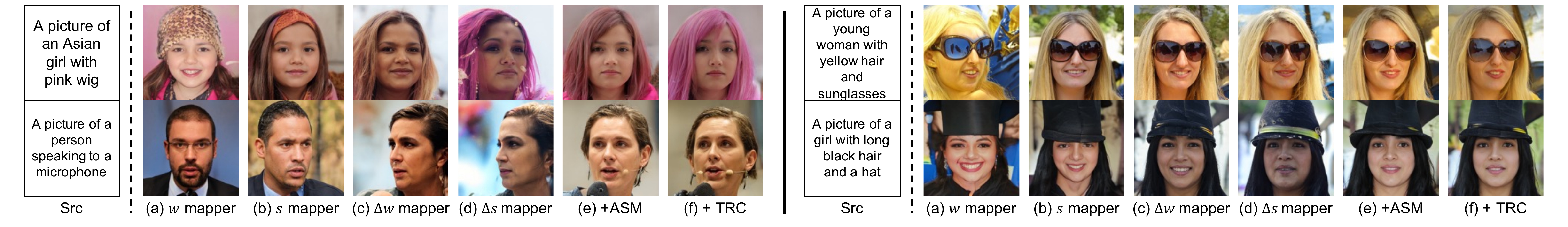}
  \caption{
    Qualitative comparison for ablation study. 
    Configurations (a)-(d) stand for different types of latent mappers.
    $\Delta s$ mapper has achieved the best performance.
    Configuration (e) introduces adaptive style mixing, which has improved the fidelity of the synthesized image.
    Configuration (f) involves temporal relative consistency, which results in a refined synthesized image.
  }
  \label{fig:ablation}
\end{figure*}

\subsection{Comparison with the State-of-the-art Methods}
As shown in \cref{tab:settings}, these SOTA methods has unaligned settings. 
To make fair comparison, the compared method should not involve extra training data, so that we have ignored AnyFace because their method depends on the dataset consists of human face images and corresponding text descriptions. 
Besides, the compared method should also support arbitrary manipulation modes.
Therefore, our major competitive methods the latent optimization mode of StyleCLIP and StyleMC despite of their much slower inference speed.

\subsubsection{Inference Speed}
According to \cref{tab:settings}, CLIP2StyleGAN is the most fast method.
This is mainly because it has pre-processed on latent directions and saved the manipulation vectors into files.
Thus, they can avoid time consumption on manipulation direction mining. 
However, the drawback is these manipulation modes are unable to be further developed.
They can only support six manipulation directions on human faces referring to their open-source code.
The inference speed of our method is in the first echelon among these SOTA methods, while ours can achieve arbitrary manipulation mode without extra data preparation during training time.

\subsubsection{Qualitative Results}
Though StyleCLIP has developed a free-generation mode by latent optimization, their generated images lack diversity and consistency to the text guidance as shown in \cref{fig:styleclip_generation}.
For the others are only proposed for manipulation, our method has achieved the best performance on text-to-image generation among these methods based on CLIP and StyleGAN.

Comparison between our method and the SOTA methods on manipulation is shown in \cref{fig:compare}.
As shown in figure, we have compared manipulation performance on four major modes.
Rather than StyleCLIP and StyleMC, our method can synthesize images with higher fidelity and more consistency with the target text prompt, while maintaining personal attributes of source images.

\subsubsection{Quantitative Results}
For quantitative comparison, we manage to evaluate these methods from two criteria: 1) personal identity; 2) consistency with the target text prompt. 
Therefore, we have employed SOTA models including ArcFace \cite{arcface} for face ID score and CLIP for text consistency score. 
100 images participated in the evaluation.
The score of face identity is computed as follows, $E_{\text{id}}$ represents the ArcFace model:
\begin{equation}
    \text{Score}_{\text{id}} = \frac{1}{n} \sum_{i=1}^{n}  
                \frac{E_{\text{id}}(I_{\text{src}})*E_{id}(I_{\text{syn}})}
                {|E_{\text{id}}(I_{\text{src}})|*|E_{id}(I_{\text{syn}}|}.
  \label{eq:id-score}
\end{equation}

Meanwhile, the CLIP score is represented by the accuracy of image classification on description text pairs. 
For example, ``a picture of a person with glasses'' and ``a picture of a person without glasses'' consist of target classes. 
Then we can compute zero-shot image classification accuracy on manipulated images.
More details of evaluation metrics are in our supplementary materials.

\begin{table}[h]
\scriptsize
  \centering
  \begin{tabular}{c|c:c|c:c|c:c}
    \toprule
    \multirow{2}{*}[-0.5ex]{\textbf{Mode}} 
    &\multicolumn{2}{c}{\textbf{StyleCLIP-optim.}} 
    &\multicolumn{2}{c}{\textbf{StyleMC}} 
    &\multicolumn{2}{c}{\textbf{Ours}} \\
    
    \cmidrule(lr){2-3}  
    \cmidrule(lr){4-5}  
    \cmidrule(lr){6-7}  
    
    
    & faceID & Acc & faceID & Acc & faceID & Acc  \\
    \hline
    glass & 0.62 & 0.36 & 0.69 & 0.30 & \textbf{0.82} &\textbf{ 0.65} \\
    \hdashline 
    smile & 0.60 & \textbf{0.99} & 0.65 & 0.98 & \textbf{0.71} & \textbf{0.99} \\
    \hdashline 
    age & 0.61 & 0.55 & 0.64 & 0.43 & \textbf{0.79} & \textbf{0.56} \\ 
    \hdashline 
    gender & 0.63 & 0.62 & 0.70 & 0.53 & \textbf{0.76} & \textbf{0.69} \\
    
    \hline
    \end{tabular}
        \caption{ 
    Quantitative comparison of three methods.
  }
  \label{tab:compare}
\end{table}

According to the evaluation results shown in \cref{tab:compare}, although other methods have utilized ArcFace and CLIP for their loss function, our method can better manipulate human face images while maintaining the face identity.

\subsection{Ablation Study}
We have made qualitative experiments for ablation study via text-to-image generation in different settings.
For text-to-image generation is sensitive to the effect of latent alignment, it can give an intuitive reflection.
As shown in \cref{fig:ablation}, the consistency with target text can be improved by using $\Delta s$ mapper.
Then, the fidelity of the synthesized image can be improved by introducing ASM.
Besides, training with TRC can further refine the synthesized image quality according to ``wig'' and ``microphone''.


\section{Conclusions}
In this paper, we aim to propose an effective and efficient method for text-driven image manipulation.
Previous works combined with CLIP and StyleGAN suffer from inference time consumption for arbitrary manipulation directions.
Therefore, we have solved this problem by aligning latents from CLIP space to $w$ and $s$ space.
Meanwhile, we also proposed adaptive style mixing (ASM) and temporal relative consistency (TRC) to improve the performance of our method.
Besides, our method can also achieve GAN inversion and text-to-image generation.
To verify our approaches, qualitative and quantitative experiments are made for comparison with SOTAs and ablation study.
Experiment results have shown that our method achieved better performance to the SOTAs.


{\small
\bibliographystyle{ieee_fullname}
\bibliography{egbib}

\begin{thebibliography}{10}\itemsep=-1pt

\bibitem{image2stylegan}
Rameen Abdal, Yipeng Qin, and Peter Wonka.
\newblock Image2stylegan: How to embed images into the stylegan latent space.
\newblock In {\em ICCV}, pages 4432--4441, 2019.

\bibitem{clip2stylegan}
Rameen Abdal, Peihao Zhu, John Femiani, Niloy~J Mitra, and Peter Wonka.
\newblock Clip2stylegan: Unsupervised extraction of stylegan edit directions.
\newblock {\em arXiv preprint arXiv:2112.05219}, 2021.

\bibitem{label4free}
Rameen Abdal, Peihao Zhu, Niloy~J Mitra, and Peter Wonka.
\newblock Labels4free: Unsupervised segmentation using stylegan.
\newblock In {\em ICCV}, pages 13970--13979, 2021.

\bibitem{restyle}
Yuval Alaluf, Or Patashnik, and Daniel Cohen-Or.
\newblock Restyle: A residual-based stylegan encoder via iterative refinement.
\newblock In {\em ICCV}, pages 6711--6720, 2021.

\bibitem{essence}
Hila Chefer, Sagie Benaim, Roni Paiss, and Lior Wolf.
\newblock Image-based clip-guided essence transfer.
\newblock {\em arXiv preprint arXiv:2110.12427}, 2021.

\bibitem{stargan}
Yunjey Choi, Minje Choi, Munyoung Kim, Jung-Woo Ha, Sunghun Kim, and Jaegul
  Choo.
\newblock {StarGAN}: Unified generative adversarial networks for multi-domain
  image-to-image translation.
\newblock In {\em CVPR}, pages 8789--8797. {IEEE}, 2018.

\bibitem{stargan2}
Yunjey Choi, Youngjung Uh, Jaejun Yoo, and Jung-Woo Ha.
\newblock {StarGAN v2}: Diverse image synthesis for multiple domains.
\newblock In {\em CVPR}, pages 8185--8194. {IEEE}, 2020.

\bibitem{jojogan}
Min~Jin Chong and David Forsyth.
\newblock {JoJoGAN}: One shot face stylization.
\newblock {\em arXiv preprint arXiv:2112.11641}, 2021.

\bibitem{vqgan-clip}
Katherine Crowson, Stella Biderman, Daniel Kornis, Dashiell Stander, Eric
  Hallahan, Louis Castricato, and Edward Raff.
\newblock Vqgan-clip: Open domain image generation and editing with natural
  language guidance.
\newblock {\em arXiv preprint arXiv:2204.08583}, 2022.

\bibitem{imagenet}
Jia Deng, Wei Dong, Richard Socher, Li-Jia Li, Kai Li, and Li Fei-Fei.
\newblock Imagenet: A large-scale hierarchical image database.
\newblock In {\em CVPR}, pages 248--255. Ieee, 2009.

\bibitem{arcface}
Jiankang Deng, Jia Guo, Niannan Xue, and Stefanos Zafeiriou.
\newblock Arcface: Additive angular margin loss for deep face recognition.
\newblock In {\em Proceedings of the IEEE/CVF conference on computer vision and
  pattern recognition}, pages 4690--4699, 2019.

\bibitem{cogview}
Ming Ding, Zhuoyi Yang, Wenyi Hong, Wendi Zheng, Chang Zhou, Da Yin, Junyang
  Lin, Xu Zou, Zhou Shao, Hongxia Yang, et~al.
\newblock Cogview: Mastering text-to-image generation via transformers.
\newblock {\em NeurIPS}, 34:19822--19835, 2021.

\bibitem{stylegan-nada}
Rinon Gal, Or Patashnik, Haggai Maron, Gal Chechik, and Daniel Cohen-Or.
\newblock Stylegan-nada: Clip-guided domain adaptation of image generators.
\newblock {\em arXiv preprint arXiv:2108.00946}, 2021.

\bibitem{clip-adapter}
Peng Gao, Shijie Geng, Renrui Zhang, Teli Ma, Rongyao Fang, Yongfeng Zhang,
  Hongsheng Li, and Yu Qiao.
\newblock Clip-adapter: Better vision-language models with feature adapters.
\newblock {\em arXiv preprint arXiv:2110.04544}, 2021.

\bibitem{gan}
Ian Goodfellow, Jean Pouget-Abadie, Mehdi Mirza, Bing Xu, David Warde-Farley,
  Sherjil Ozair, Aaron Courville, and Yoshua Bengio.
\newblock Generative adversarial nets.
\newblock In {\em NeurIPS}, pages 2672--2680, 2014.

\bibitem{ddpm}
Jonathan Ho, Ajay Jain, and Pieter Abbeel.
\newblock Denoising diffusion probabilistic models.
\newblock {\em NeurIPS}, 33:6840--6851, 2020.

\bibitem{stylegan3}
Tero Karras, Miika Aittala, Samuli Laine, Erik H{\"a}rk{\"o}nen, Janne
  Hellsten, Jaakko Lehtinen, and Timo Aila.
\newblock Alias-free generative adversarial networks.
\newblock {\em NeurIPS}, 34:852--863, 2021.

\bibitem{stylegan}
Tero Karras, Samuli Laine, and Timo Aila.
\newblock A style-based generator architecture for generative adversarial
  networks.
\newblock In {\em CVPR}, pages 4401--4410. {IEEE}, 2019.

\bibitem{stylegan2}
Tero Karras, Samuli Laine, Miika Aittala, Janne Hellsten, Jaakko Lehtinen, and
  Timo Aila.
\newblock Analyzing and improving the image quality of {StyleGAN}.
\newblock In {\em CVPR}, pages 8110--8119. {IEEE}, 2020.

\bibitem{diffusionclip}
Gwanghyun Kim and Jong~Chul Ye.
\newblock Diffusionclip: Text-guided image manipulation using diffusion models.
\newblock 2021.

\bibitem{adam}
Diederik~P Kingma and Jimmy Ba.
\newblock {Adam}: A method for stochastic optimization.
\newblock In {\em ICLR}, 2015.

\bibitem{stylemc}
Umut Kocasari, Alara Dirik, Mert Tiftikci, and Pinar Yanardag.
\newblock Stylemc: Multi-channel based fast text-guided image generation and
  manipulation.
\newblock In {\em WACV}, pages 895--904, 2022.

\bibitem{glide}
Alex Nichol, Prafulla Dhariwal, Aditya Ramesh, Pranav Shyam, Pamela Mishkin,
  Bob McGrew, Ilya Sutskever, and Mark Chen.
\newblock Glide: Towards photorealistic image generation and editing with
  text-guided diffusion models.
\newblock {\em arXiv preprint arXiv:2112.10741}, 2021.

\bibitem{styleclip}
Or Patashnik, Zongze Wu, Eli Shechtman, Daniel Cohen-Or, and Dani Lischinski.
\newblock Styleclip: Text-driven manipulation of stylegan imagery.
\newblock In {\em ICCV}, pages 2085--2094, 2021.

\bibitem{clip2latent}
Justin~NM Pinkney and Chuan Li.
\newblock clip2latent: Text driven sampling of a pre-trained stylegan using
  denoising diffusion and clip.
\newblock {\em BMVC}, 2022.

\bibitem{clip}
Alec Radford, Jong~Wook Kim, Chris Hallacy, Aditya Ramesh, Gabriel Goh,
  Sandhini Agarwal, Girish Sastry, Amanda Askell, Pamela Mishkin, Jack Clark,
  et~al.
\newblock Learning transferable visual models from natural language
  supervision.
\newblock In {\em ICML}, pages 8748--8763. PMLR, 2021.

\bibitem{dalle2}
Aditya Ramesh, Prafulla Dhariwal, Alex Nichol, Casey Chu, and Mark Chen.
\newblock Hierarchical text-conditional image generation with clip latents.
\newblock {\em arXiv preprint arXiv:2204.06125}, 2022.

\bibitem{dalle}
Aditya Ramesh, Mikhail Pavlov, Gabriel Goh, Scott Gray, Chelsea Voss, Alec
  Radford, Mark Chen, and Ilya Sutskever.
\newblock Zero-shot text-to-image generation.
\newblock In {\em ICML}, pages 8821--8831. PMLR, 2021.

\bibitem{denseclip}
Yongming Rao, Wenliang Zhao, Guangyi Chen, Yansong Tang, Zheng Zhu, Guan Huang,
  Jie Zhou, and Jiwen Lu.
\newblock Denseclip: Language-guided dense prediction with context-aware
  prompting.
\newblock In {\em CVPR}, pages 18082--18091, 2022.

\bibitem{psp}
Elad Richardson, Yuval Alaluf, Or Patashnik, Yotam Nitzan, Yaniv Azar, Stav
  Shapiro, and Daniel Cohen-Or.
\newblock Encoding in style: a stylegan encoder for image-to-image translation.
\newblock In {\em CVPR}, pages 2287--2296, 2021.

\bibitem{stylegan-rosi}
rosinality.
\newblock Implementation of analyzing and improving the image quality of
  stylegan in pytorch.
\newblock \url{https://github.com/rosinality/stylegan2-pytorch}, 2019.

\bibitem{dreambooth}
Nataniel Ruiz, Yuanzhen Li, Varun Jampani, Yael Pritch, Michael Rubinstein, and
  Kfir Aberman.
\newblock Dreambooth: Fine tuning text-to-image diffusion models for
  subject-driven generation.
\newblock {\em arXiv preprint arXiv:2208.12242}, 2022.

\bibitem{imagen}
Chitwan Saharia, William Chan, Saurabh Saxena, Lala Li, Jay Whang, Emily
  Denton, Seyed Kamyar~Seyed Ghasemipour, Burcu~Karagol Ayan, S~Sara Mahdavi,
  Rapha~Gontijo Lopes, et~al.
\newblock Photorealistic text-to-image diffusion models with deep language
  understanding.
\newblock {\em arXiv preprint arXiv:2205.11487}, 2022.

\bibitem{styleganxl}
Axel Sauer, Katja Schwarz, and Andreas Geiger.
\newblock Stylegan-xl: Scaling stylegan to large diverse datasets.
\newblock In {\em SIGGRAPH}, pages 1--10, 2022.

\bibitem{interfacegan}
Yujun Shen, Jinjin Gu, Xiaoou Tang, and Bolei Zhou.
\newblock Interpreting the latent space of gans for semantic face editing.
\newblock In {\em CVPR}, pages 9243--9252, 2020.

\bibitem{flava}
Amanpreet Singh, Ronghang Hu, Vedanuj Goswami, Guillaume Couairon, Wojciech
  Galuba, Marcus Rohrbach, and Douwe Kiela.
\newblock Flava: A foundational language and vision alignment model.
\newblock In {\em CVPR}, pages 15638--15650. IEEE, June 2022.

\bibitem{diffusion}
Jascha Sohl-Dickstein, Eric Weiss, Niru Maheswaranathan, and Surya Ganguli.
\newblock Deep unsupervised learning using nonequilibrium thermodynamics.
\newblock In {\em ICML}, pages 2256--2265. PMLR, 2015.

\bibitem{diffusion2}
Yang Song and Stefano Ermon.
\newblock Generative modeling by estimating gradients of the data distribution.
\newblock {\em NeurIPS}, 32, 2019.

\bibitem{anyface}
Jianxin Sun, Qiyao Deng, Qi Li, Muyi Sun, Min Ren, and Zhenan Sun.
\newblock Anyface: Free-style text-to-face synthesis and manipulation.
\newblock In {\em CVPR}, pages 18687--18696, 2022.

\bibitem{e4e}
Omer Tov, Yuval Alaluf, Yotam Nitzan, Or Patashnik, and Daniel Cohen-Or.
\newblock Designing an encoder for stylegan image manipulation.
\newblock {\em ACM Transactions on Graphics (TOG)}, 40(4):1--14, 2021.

\bibitem{contraclip}
Christos Tzelepis, James Oldfield, Georgios Tzimiropoulos, and Ioannis Patras.
\newblock Contraclip: Interpretable gan generation driven by pairs of
  contrasting sentences.
\newblock {\em arXiv preprint arXiv:2206.02104}, 2022.

\bibitem{transformer}
Ashish Vaswani, Noam Shazeer, Niki Parmar, Jakob Uszkoreit, Llion Jones,
  Aidan~N Gomez, {\L}ukasz Kaiser, and Illia Polosukhin.
\newblock Attention is all you need.
\newblock In {\em NeurIPS}, pages 5998--6008, 2017.

\bibitem{clip-gen}
Zihao Wang, Wei Liu, Qian He, Xinglong Wu, and Zili Yi.
\newblock Clip-gen: Language-free training of a text-to-image generator with
  clip.
\newblock {\em arXiv preprint arXiv:2203.00386}, 2022.

\bibitem{hairclip}
Tianyi Wei, Dongdong Chen, Wenbo Zhou, Jing Liao, Zhentao Tan, Lu Yuan, Weiming
  Zhang, and Nenghai Yu.
\newblock Hairclip: Design your hair by text and reference image.
\newblock In {\em CVPR}, pages 18072--18081, 2022.

\bibitem{lsun}
Fisher Yu, Ari Seff, Yinda Zhang, Shuran Song, Thomas Funkhouser, and Jianxiong
  Xiao.
\newblock Lsun: Construction of a large-scale image dataset using deep learning
  with humans in the loop.
\newblock {\em arXiv preprint arXiv:1506.03365}, 2015.

\bibitem{counterfactual}
Yingchen Yu, Fangneng Zhan, Rongliang Wu, Jiahui Zhang, Shijian Lu, Miaomiao
  Cui, Xuansong Xie, Xian-Sheng Hua, and Chunyan Miao.
\newblock Towards counterfactual image manipulation via clip.
\newblock {\em arXiv preprint arXiv:2207.02812}, 2022.

\bibitem{cocop}
Kaiyang Zhou, Jingkang Yang, Chen~Change Loy, and Ziwei Liu.
\newblock Conditional prompt learning for vision-language models.
\newblock In {\em CVPR}, pages 16816--16825. IEEE, June 2022.

\bibitem{coop}
Kaiyang Zhou, Jingkang Yang, Chen~Change Loy, and Ziwei Liu.
\newblock Learning to prompt for vision-language models.
\newblock {\em IJCV}, pages 1--12, 2022.

\end{thebibliography}
}

\newpage

\appendix





\flushleft{\textbf{\huge{Appendix}}}

\section{Overview}
In this supplementary material, we enumerate the text pairs for quantitative analysis and present more experimental results.
\begin{itemize}
    \item Enumeration of text pairs for quantitative analysis mentioned in the main paper.

    \item Experiments results of StyleGAN2 \cite{stylegan2}, which is pre-trained on cat, car, church, horse datasets from LSUN \cite{lsun}.
    
    \item Experiments results of StyleGAN3 \cite{stylegan3}, which is pre-trained on animal face dataset AFHQ \cite{stargan,stargan2}.
    
    \item Experiments results of StyleGAN-XL \cite{styleganxl}, which is pre-trained on ImageNet \cite{imagenet}.
    

\end{itemize}

According to experiment results demonstrated in the following pages, our method can not only achieve attribute or shape manipulation on target object, but also change its motion. Beyond that, we can even change the background and transfer the image to grayscale via text prompt.


\newpage
\section{Text for Quantitative Analysis}

\begin{table}[h]
\scriptsize
  \centering
  \begin{tabular}{c|c:c}
    \toprule
    
    \textbf{Mode}
    &\textbf{Positive}
    &\textbf{Negative} \\

    \midrule

    glass & ``a picture of a person with glasses'' & ``a picture of a person without glasses'' \\
    \hdashline

    smile & ``a picture of a smiling person'' & ``a picture of a serious person'' \\
    \hdashline

    age & ``a picture of an old person'' & ``a picture of a young person'' \\
    \hdashline

    gender & ``a picture of a man'' & ``a picture of a woman'' \\
    \hdashline

    
    
    \hline
    \end{tabular}
        \caption{ 
    Texts pairs for quantitative comparison.
  }
  \label{tab:texts}
\end{table}

\section{Other experiment results}

\begin{figure*}[h]
  \centering
  \includegraphics[width=\linewidth]{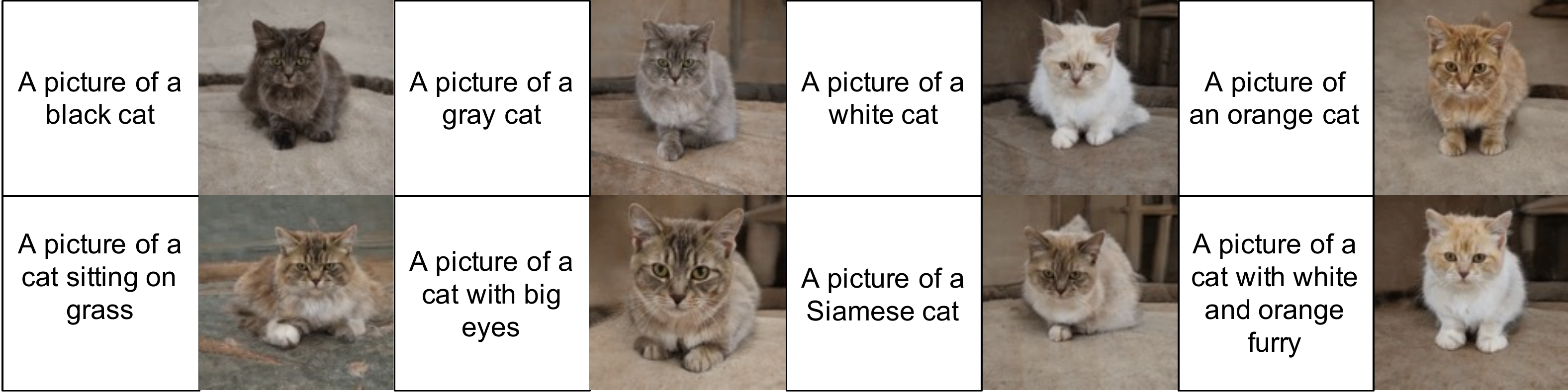}
  \caption{
    Demonstration of text-to-image generation results on LSUN cat.
  }
  \label{fig:cat1}
\end{figure*}

\newpage

\begin{figure*}[t]
  \centering
  \includegraphics[width=\linewidth]{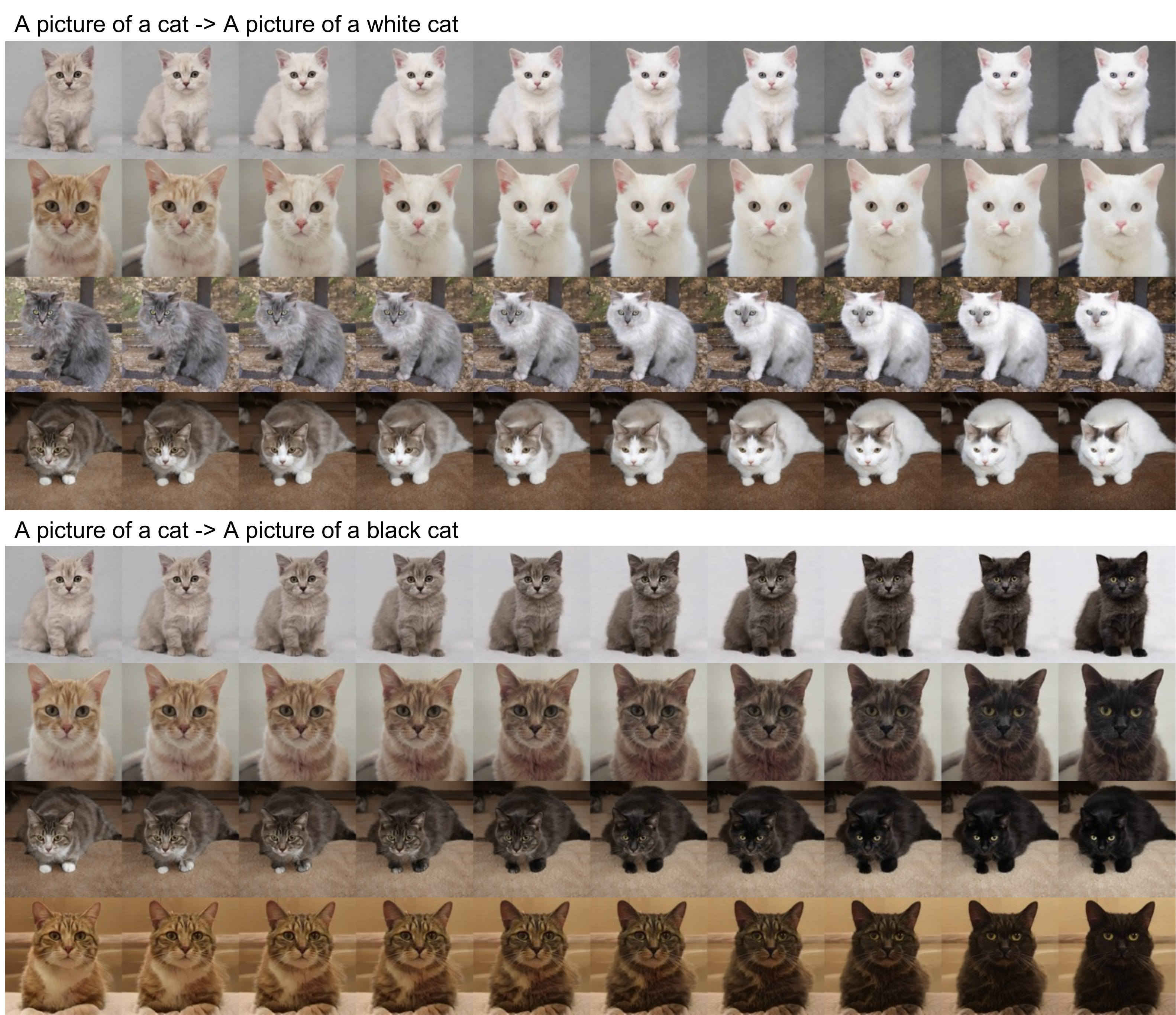}
  \caption{
    Demonstration of text-driven image manipulation results on LSUN cat part 1.
  }
  \label{fig:cat2}
\end{figure*}

\begin{figure*}[t]
  \centering
  \includegraphics[width=\linewidth]{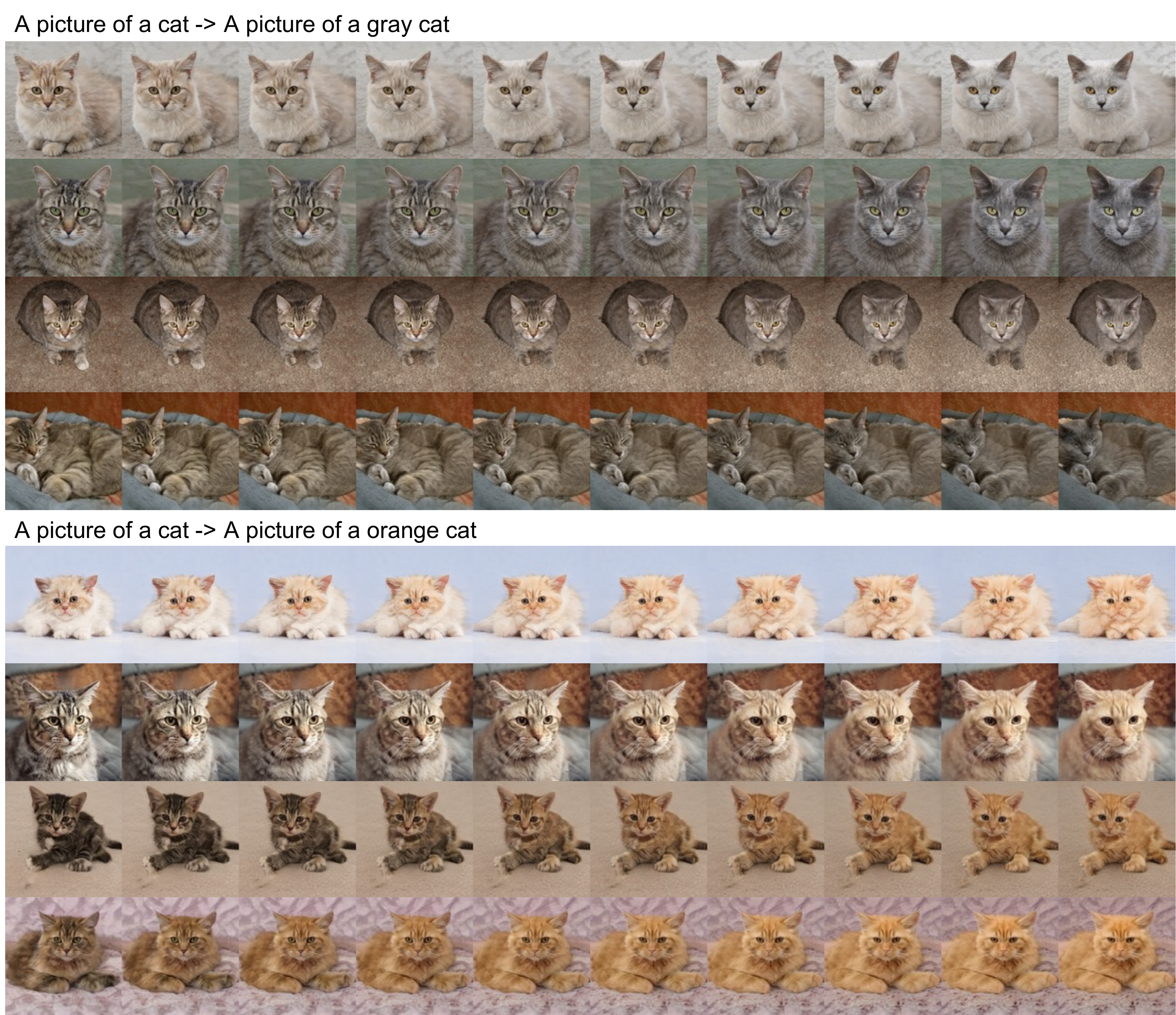}
  \caption{
    Demonstration of text-driven image manipulation results on LSUN cat part 2.
  }
  \label{fig:cat3}
\end{figure*}


\begin{figure*}[h]
  \centering
  \includegraphics[width=\linewidth]{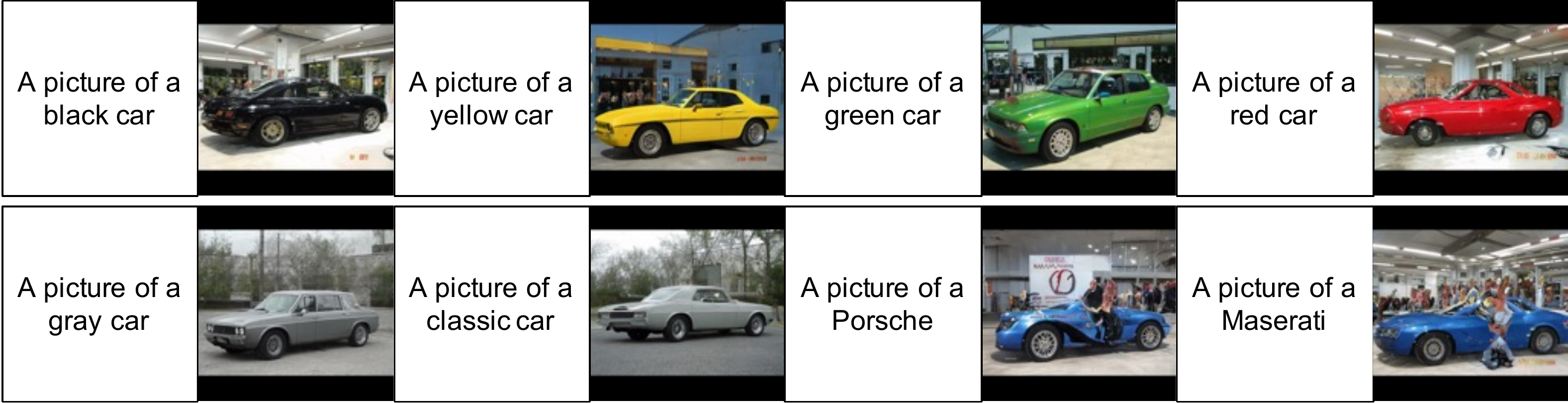}
  \caption{
    Demonstration of text-to-image generation results on LSUN car.
  }
  \label{fig:car1}
\end{figure*}

\begin{figure*}[h]
  \centering
  \includegraphics[width=\linewidth]{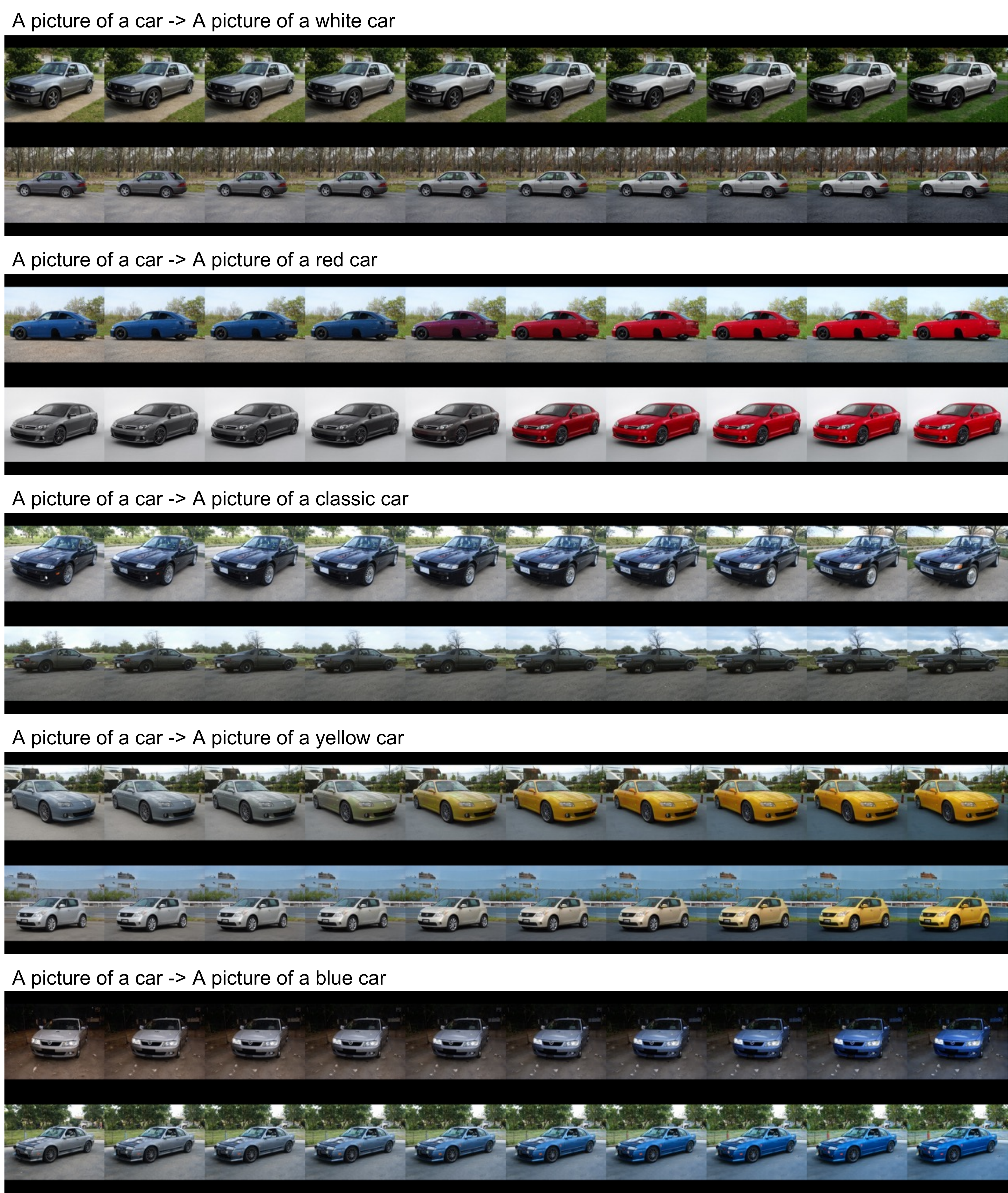}
  \caption{
    Demonstration of text-driven image manipulation results on LSUN car.
  }
  \label{fig:car2}
\end{figure*}

\begin{figure*}[h]
  \centering
  \includegraphics[width=\linewidth]{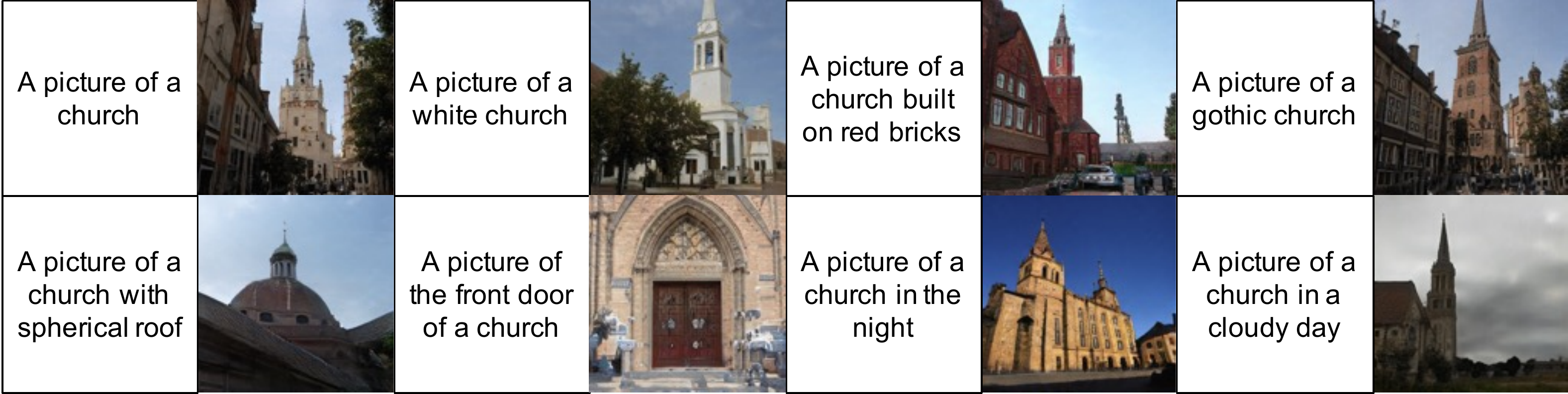}
  \caption{
    Demonstration of text-to-image generation results on LSUN church.
  }
  \label{fig:chruch1}
\end{figure*}

\begin{figure*}[h]
  \centering
  \includegraphics[width=\linewidth]{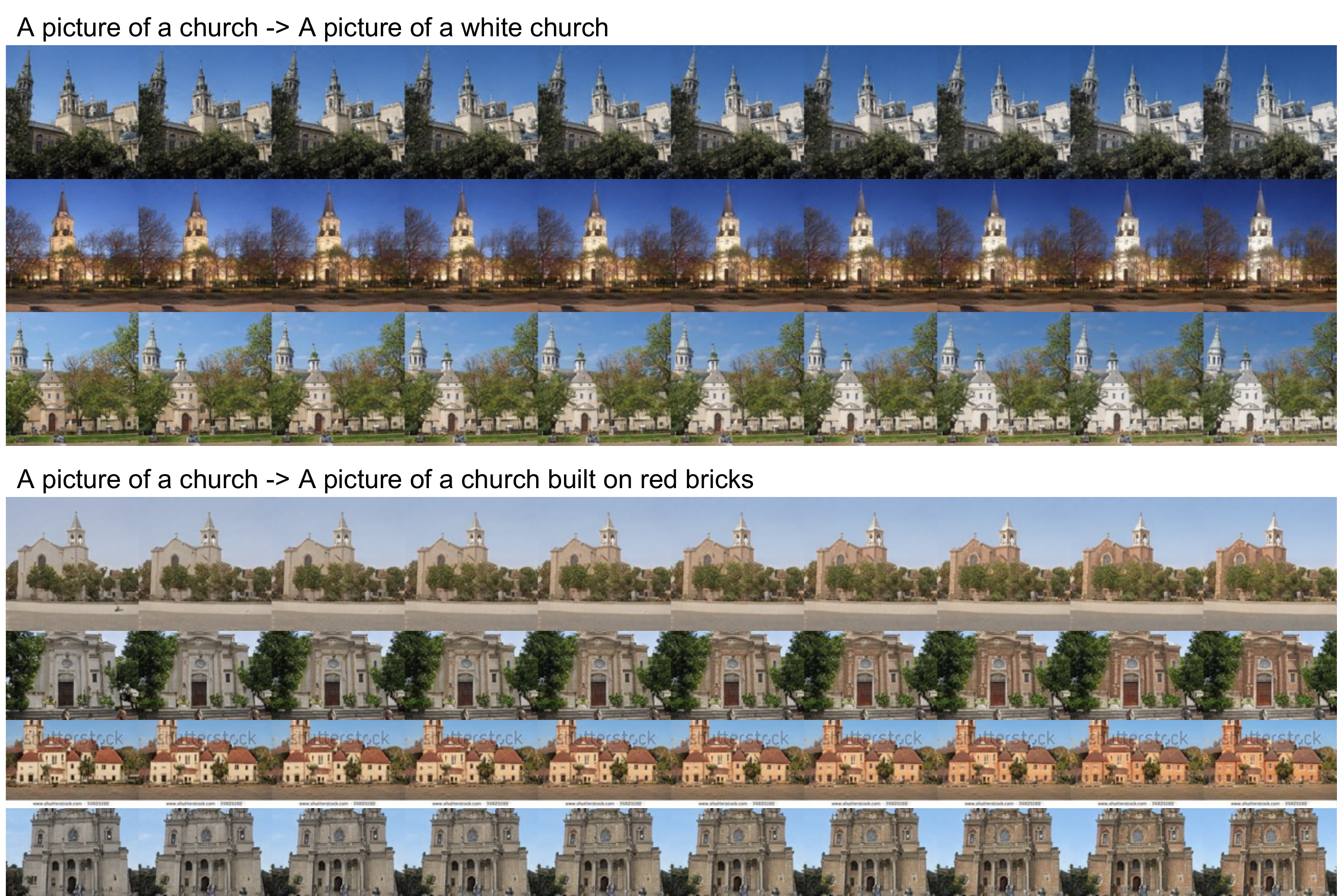}
  \caption{
    Demonstration of text-driven image manipulation results on LSUN church part 1.
  }
  \label{fig:church2}
\end{figure*}

\begin{figure*}[h]
  \centering
  \includegraphics[width=\linewidth]{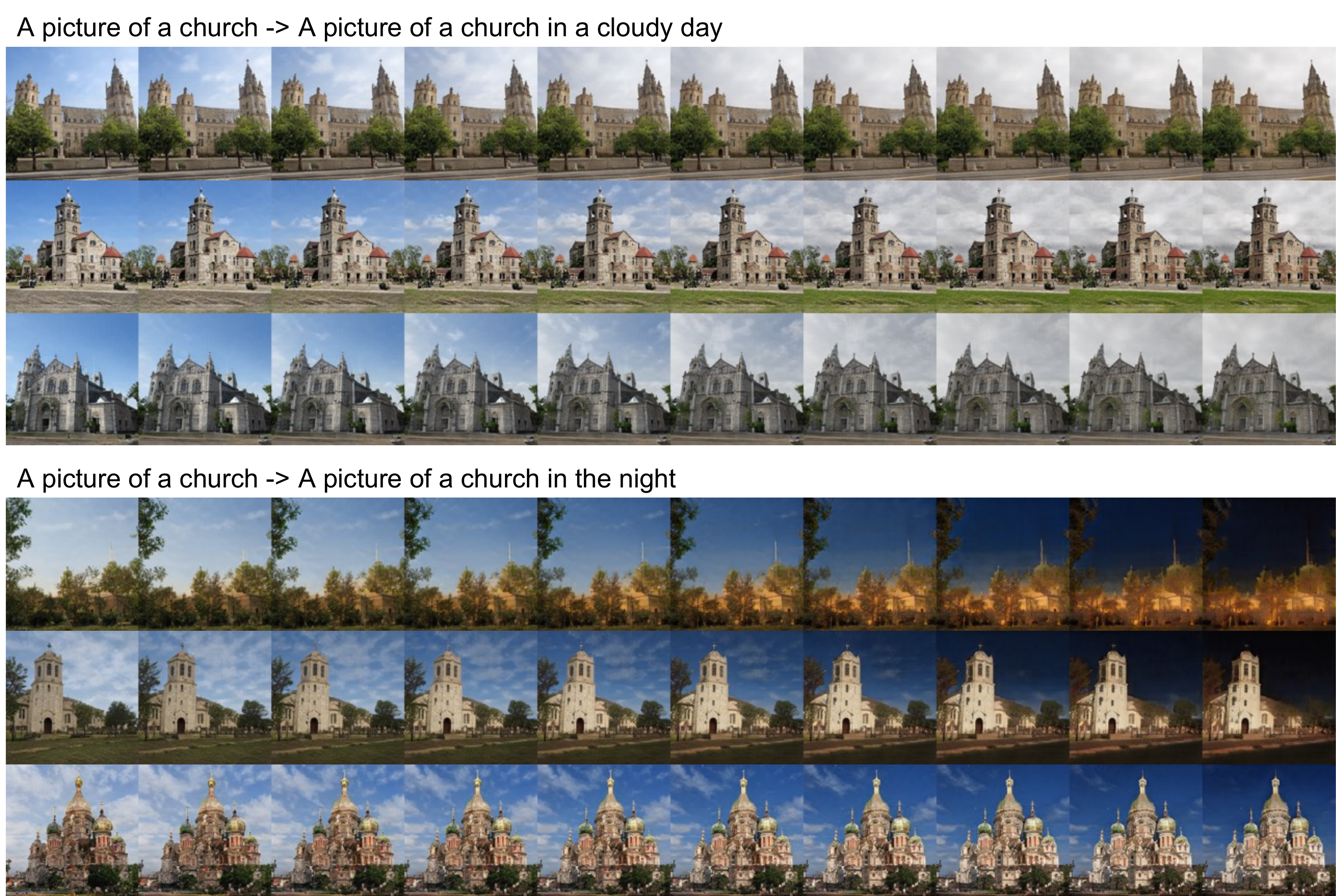}
  \caption{
    Demonstration of text-driven image manipulation results on LSUN church part 2.
  }
  \label{fig:church3}
\end{figure*}

\begin{figure*}[h]
  \centering
  \includegraphics[width=\linewidth]{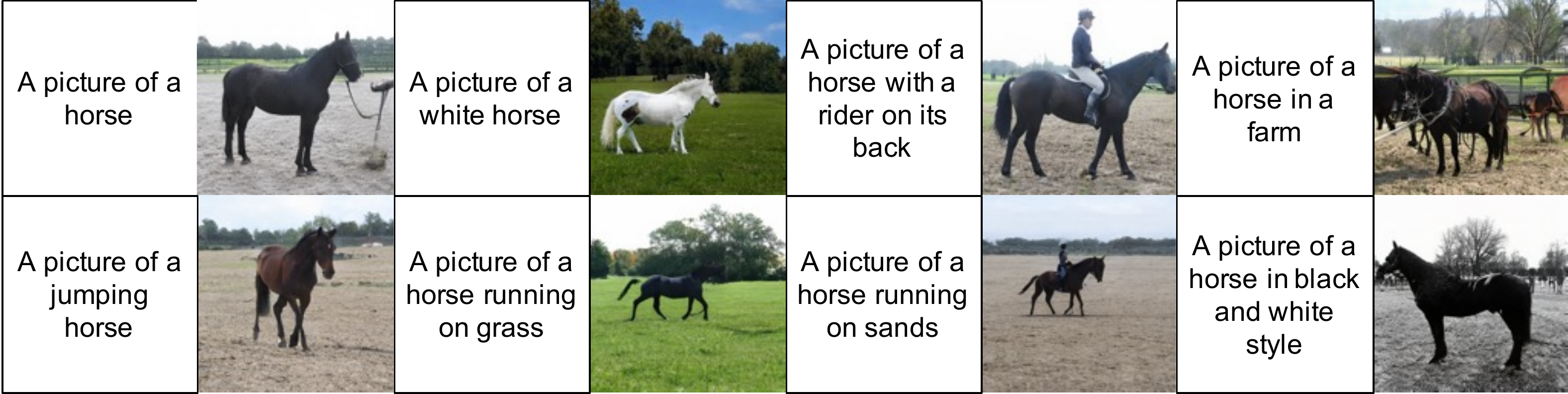}
  \caption{
    Demonstration of text-to-image generation results on LSUN hosrse.
  }
  \label{fig:horse1}
\end{figure*}



\begin{figure*}[h]
  \centering
  \includegraphics[width=\linewidth]{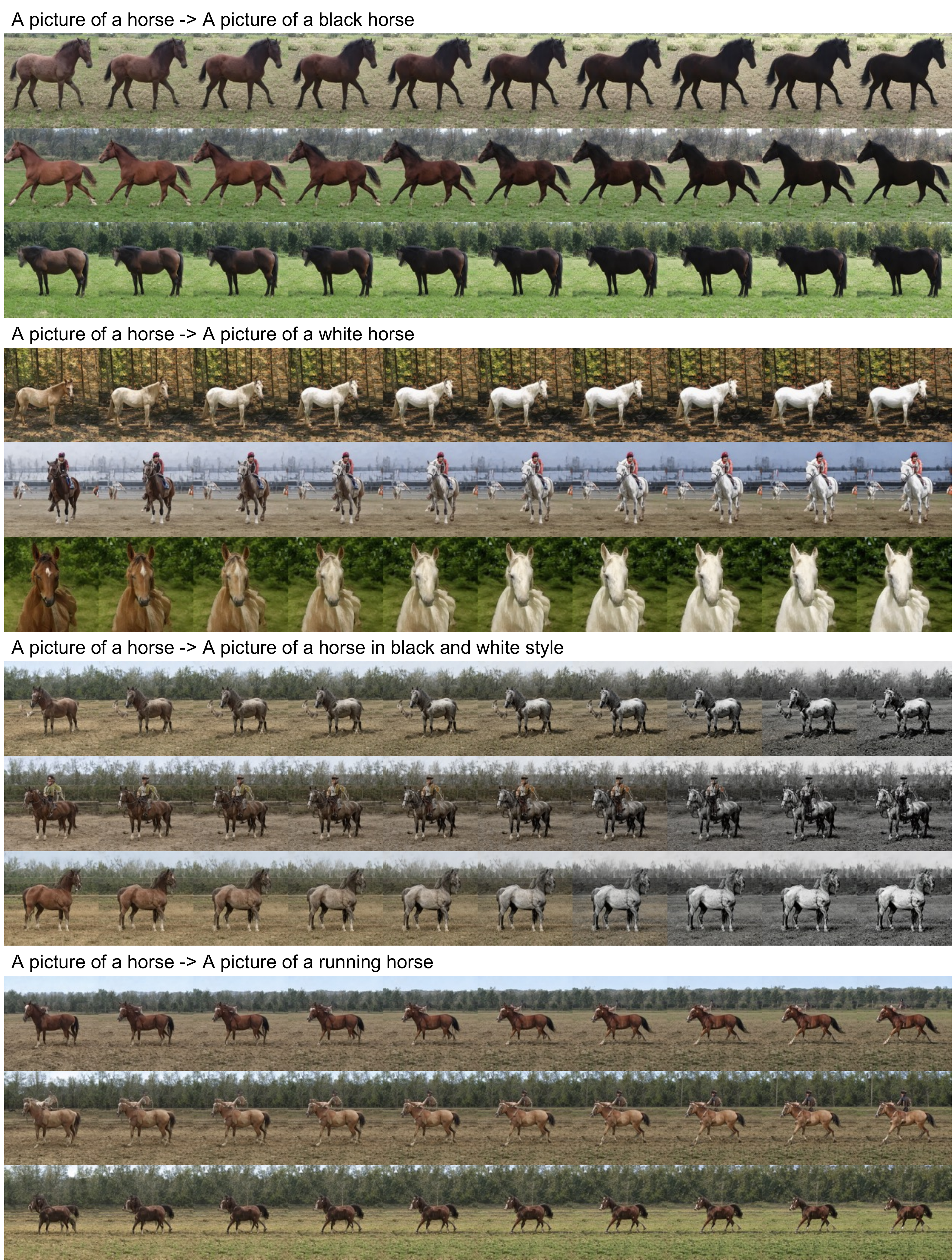}
  \caption{
    Demonstration of text-driven image manipulation results on LSUN horse.
  }
  \label{fig:horse4}
\end{figure*}

\begin{figure*}[h]
  \centering
  \includegraphics[width=\linewidth]{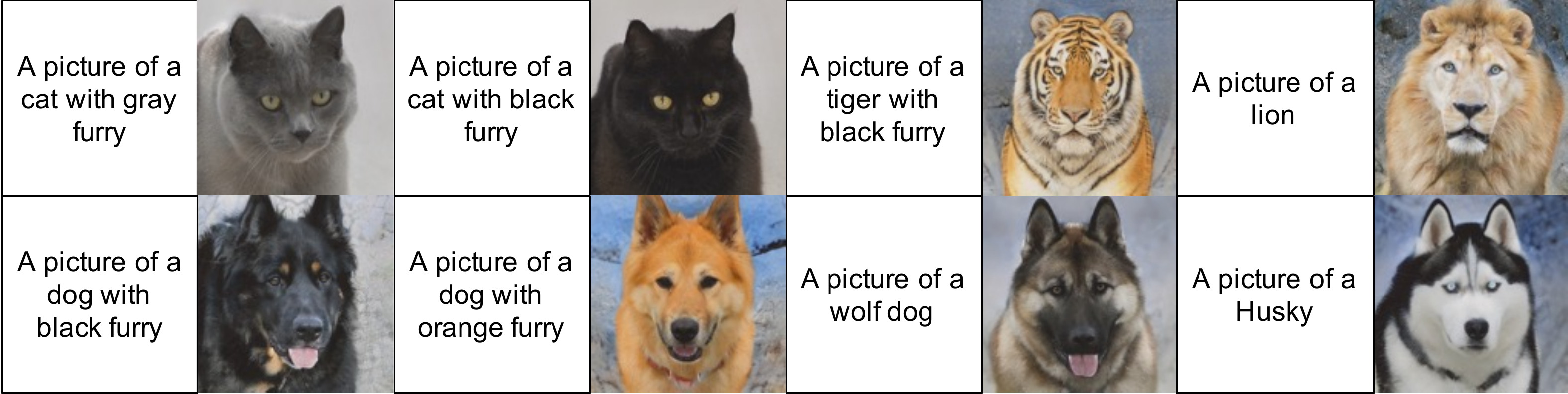}
  \caption{
    Demonstration of text-to-image generation results with StyleGAN3 pre-trained on AFHQ.
  }
  \label{fig:afhq1}
\end{figure*}

\begin{figure*}[h]
  \centering
  \includegraphics[width=\linewidth]{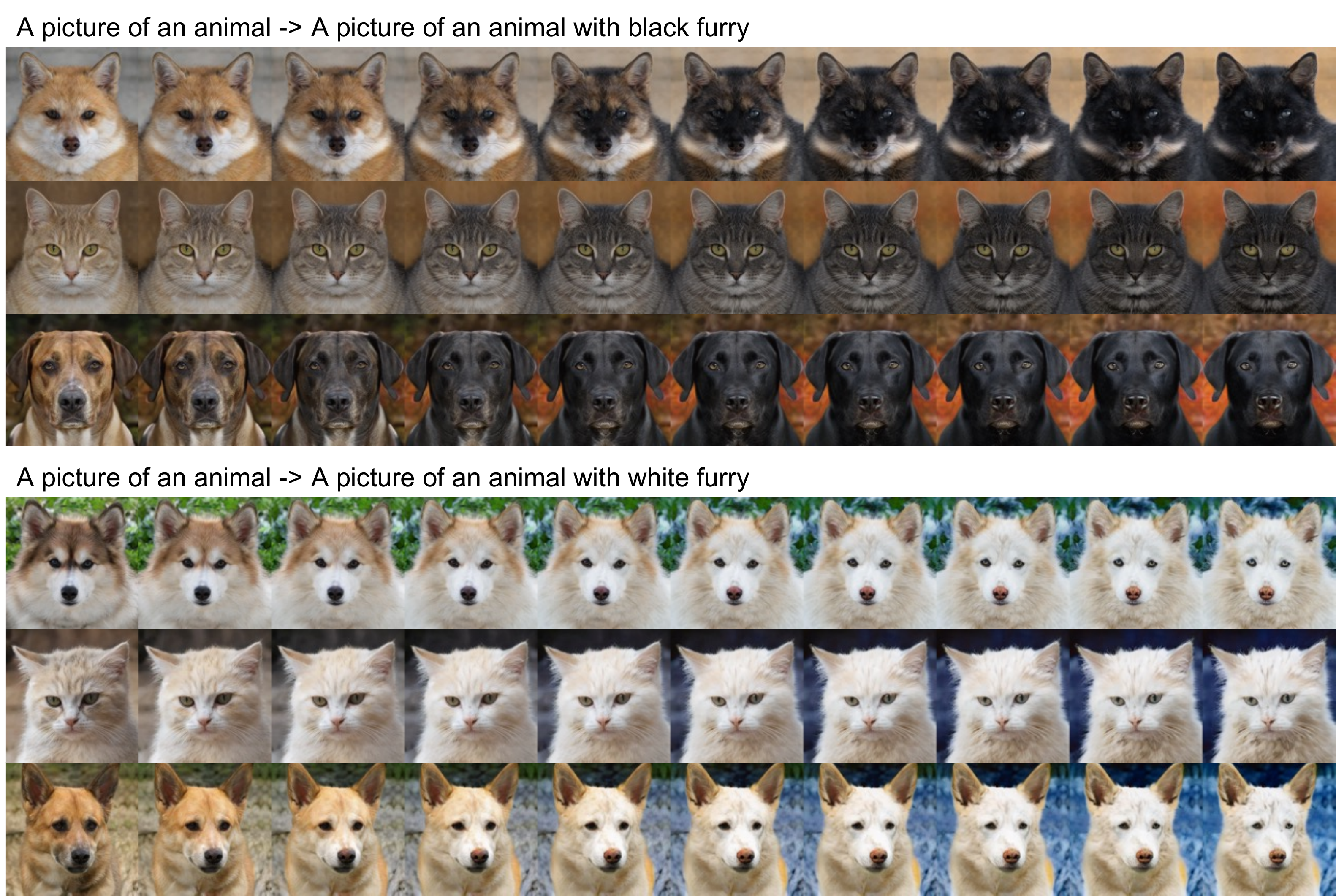}
  \caption{
    Demonstration of text-driven manipulation results on AFHQ part 1.
  }
  \label{fig:afhq2}
\end{figure*}

\begin{figure*}[h]
  \centering
  \includegraphics[width=\linewidth]{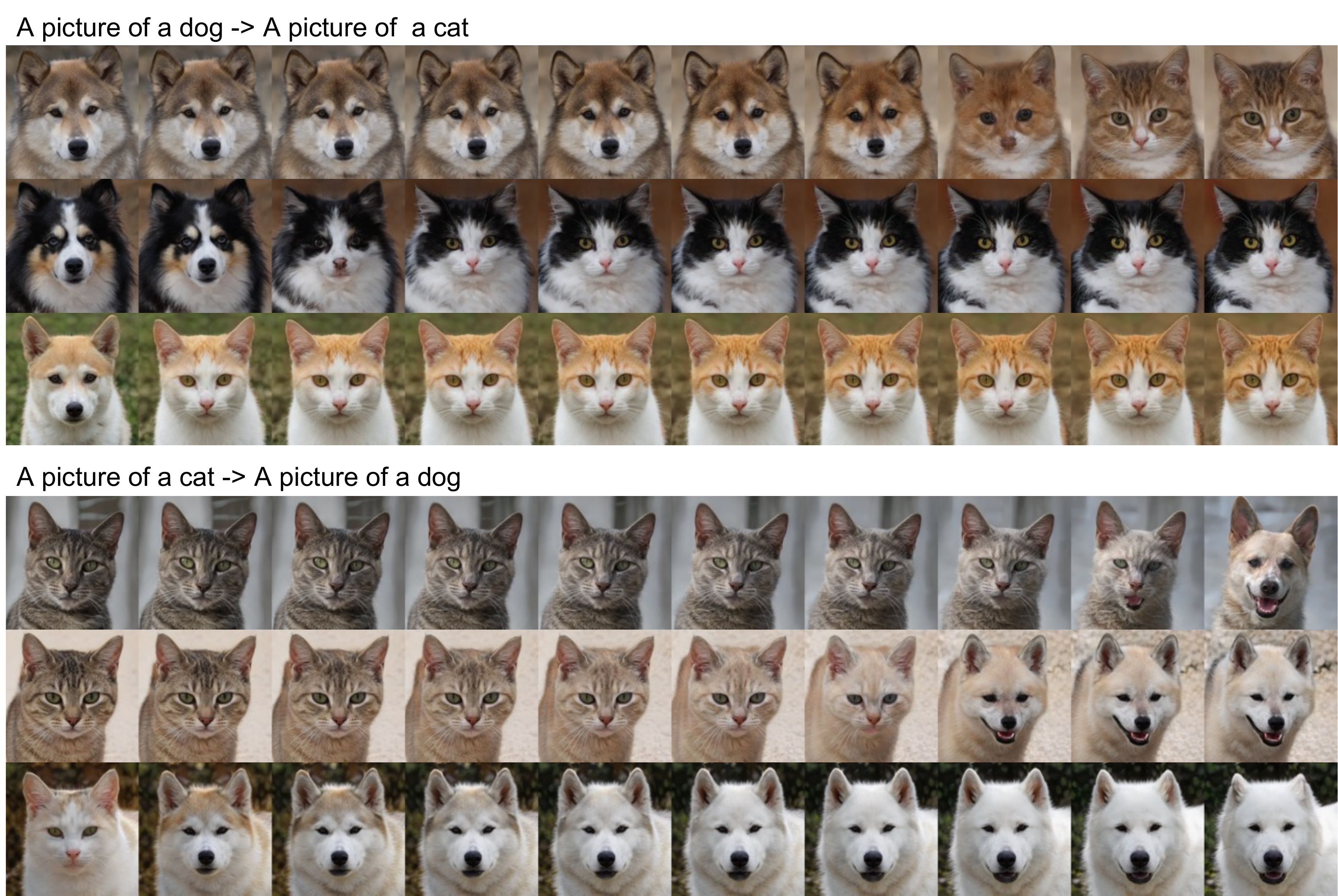}
  \caption{
    Demonstration of text-driven manipulation results on AFHQ part 2.
  }
  \label{fig:afhq3}
\end{figure*}

\begin{figure*}[h]
  \centering
  \includegraphics[width=\linewidth]{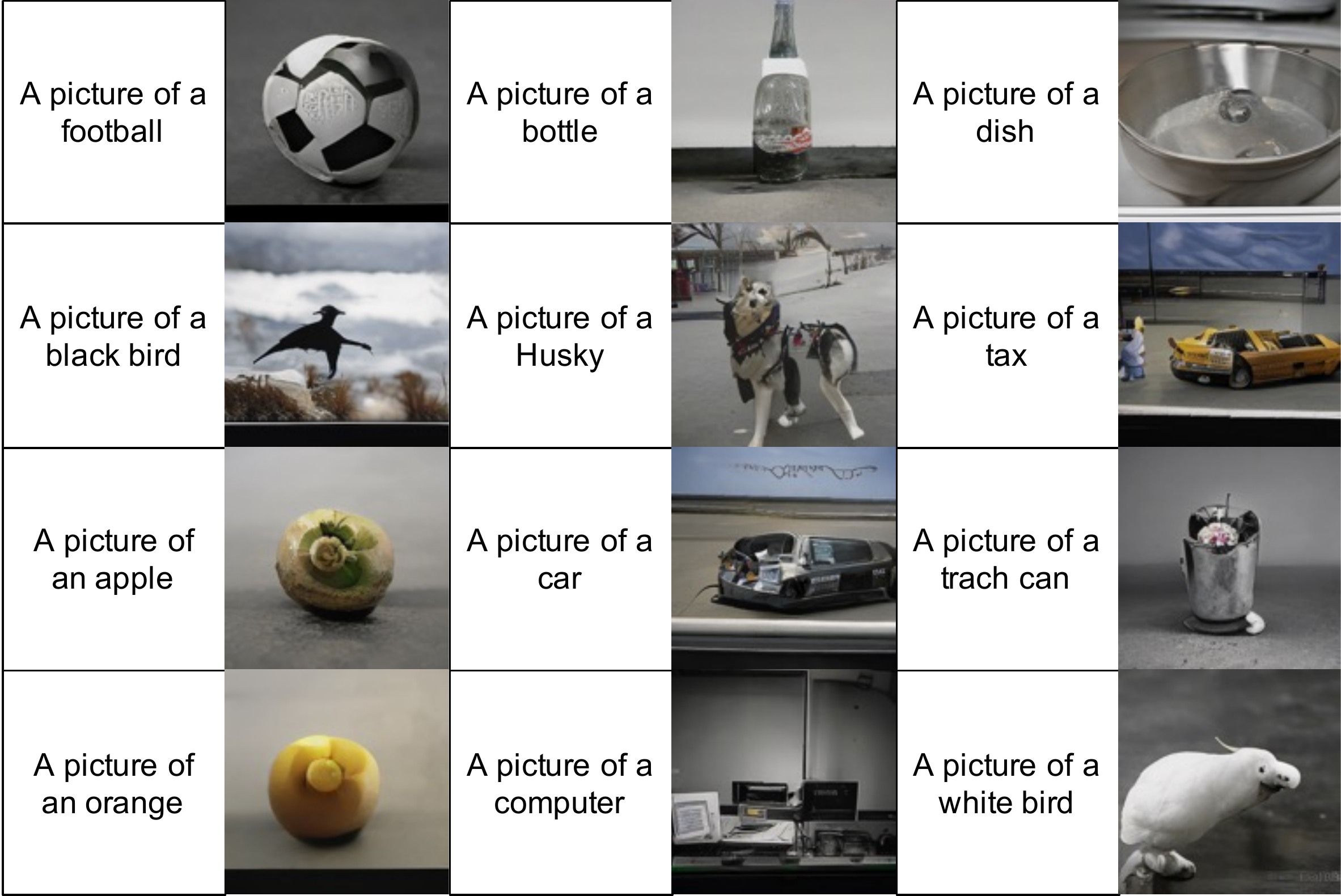}
  \caption{
    Demonstration of text-to-image generation results with StyleGAN-XL pre-trained on ImageNet.
    We have not made manipulation experiments on this dataset due to its complicated object classes and the limited generative ability of StyleGAN-XL.
  }
  \label{fig:imagenet1}
\end{figure*}

\end{document}